\documentclass[journal]{IEEEtran}
\usepackage{amsmath,amsfonts}
\usepackage[ruled,vlined,linesnumbered]{algorithm2e}
\usepackage{algorithmic}
\usepackage{array}
\usepackage[caption=false,font=normalsize,labelfont=sf,textfont=sf]{subfig}
\usepackage{textcomp}
\usepackage{stfloats}
\usepackage{url}
\usepackage{verbatim}
\usepackage{graphicx}
\usepackage{multirow}
\usepackage{booktabs}
\usepackage{cite}
\usepackage{color}
\usepackage{hyperref}
\begin{document}

\title{Learning to Denoise Biomedical Knowledge Graph for Robust Molecular Interaction Prediction}

% author, tengfei ma
\author{Tengfei Ma,
        Yujie Chen, Wen Tao, Dashun Zheng, Xuan Lin, Patrick Cheong-Iao Pang, Yiping Liu, Yijun Wang, Longyue Wang,
        Bosheng Song,
        Xiangxiang Zeng$^*$,~\IEEEmembership{Senior Member,~IEEE},
        and Philip S. Yu,~\IEEEmembership{Fellow,~IEEE}
        % <-this % stops a space

\thanks{Tengfei Ma, Yujie Chen, Wen Tao, Yiping Liu, Yijun Wang, Bosheng Song, and Xiangxiang Zeng are with the College of Computer Science and Electronic Engineering, Hunan University, China. (E-mail: \{tfma, yjchen, taowen, ypliu, wyjun, boshengsong, xzeng\}@hnu.edu.cn).}
\thanks{Dashun Zheng and Patrick Cheong-Iao Pang are with the Faculty of Applied Sciences, Macao Polytechnic University, Macao, China. (E-mail: p2212871@mpu.edu.mo; mail@patrickpang.net).}
\thanks{Xuan Lin is with the College of Computer Science, Xiangtan University, China. (E-mail: jack\_lin@xtu.edu.cn).}
\thanks{Longyue Wang is with the Tencent AI Lab, Shenzhen, China. (E-mail: vinnylywang@tencent.com).}
\thanks{Philip S. Yu is with the Department of Computer Science, University of
Illinois at Chicago, Chicago, IL 60607-7053 USA. (E-mail: psyu@uic.edu).}
\thanks{\textit{$^*$Corresponding author: Xiangxiang Zeng}.}
}

\markboth{Journal of \LaTeX\ Class Files,~Vol.~14, No.~8, August~2021}%
{Shell \MakeLowercase{\textit{et al.}}: A Sample Article Using IEEEtran.cls for IEEE Journals}


\maketitle

\begin{abstract}
Molecular interaction prediction plays a crucial role in forecasting unknown interactions between molecules, such as drug-target interaction (DTI) and drug-drug interaction (DDI), which are essential in the field of drug discovery and therapeutics. Although previous prediction methods have yielded promising results by leveraging the rich semantics and topological structure of biomedical knowledge graphs (KGs), they have primarily focused on enhancing predictive performance without addressing the presence of inevitable noise and inconsistent semantics. This limitation has hindered the advancement of KG-based prediction methods. To address this limitation, we propose BioKDN (\textbf{Bio}medical \textbf{K}nowledge Graph \textbf{D}enoising \textbf{N}etwork) for robust molecular interaction prediction. BioKDN refines the reliable structure of local subgraphs by denoising noisy links in a learnable manner, providing a general module for extracting task-relevant interactions. To enhance the reliability of the refined structure, 
BioKDN maintains consistent and robust semantics by smoothing relations around the target interaction. By maximizing the mutual information between reliable structure and smoothed relations, BioKDN emphasizes informative semantics to enable precise predictions. Experimental results on real-world datasets show that BioKDN surpasses state-of-the-art models in DTI and DDI prediction tasks, confirming the effectiveness and robustness of BioKDN in denoising unreliable interactions within contaminated KGs. Code is available at \url{https://github.com/xiaomingaaa/BioKDN}.
\end{abstract}


% \section{Introduction}
% \IEEEPARstart{T}{his} file is intended to serve as a ``sample article file''
\section{Introduction}

% Identifying missing molecular interactions 
The prediction of molecular interactions, including drug-target interaction (DTI) prediction~\cite{ye2021unified} and drug-drug interaction (DDI) prediction~\cite{lin2020kgnn,yu2021sumgnn,lin2023comprehensive}, is pivotal to drug discovery and therapeutics.
The success of interaction prediction based on knowledge graphs (KGs) in \textcolor{black}{social networks~\cite{he2020constructing,liang2024survey}} and recommendations~\cite{wang2019explainable,yang2022knowledge,guo2021trust} encourages researchers to develop various KG-based computational methods to accelerate drug development~\cite{ma2022kg,wu2023megacare,wu2023survey,zhang2024pre}. \textcolor{black}{The biomedical KGs contain a large number of entities with specific patterns and semantics (e.g., the associated pathways between drugs and targets). The interaction patterns between biomedical entities can enhance the prediction of biological connections among various entities~\cite{ma2022kg}.}
However, accurately recognizing the unknown interactions between various molecular entities with computational models remains challenging.

\begin{figure}[t]
\centering
\includegraphics[width=0.95\columnwidth]{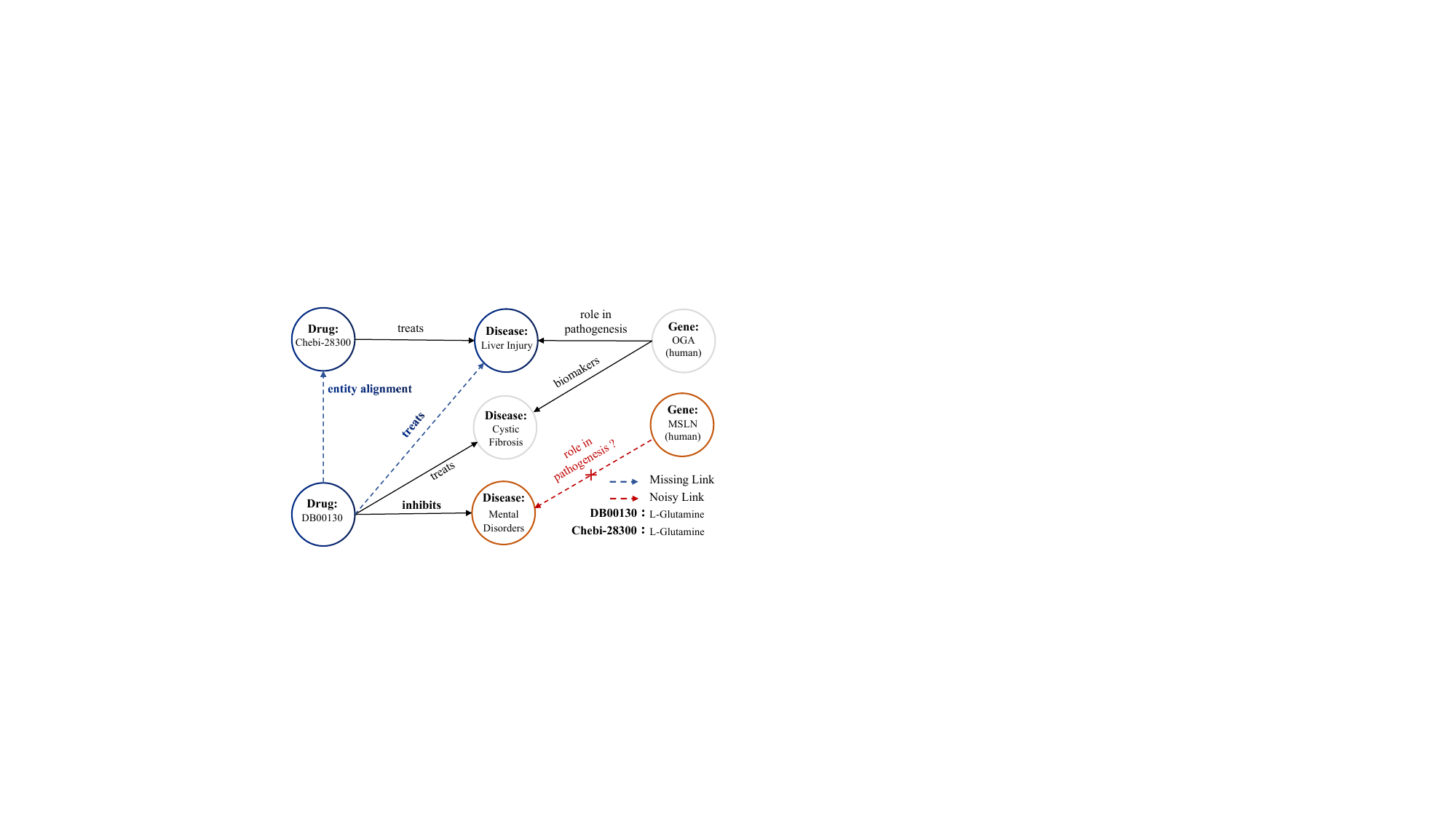} % Reduce the Fig. size so that it is slightly narrower than the column. Don't use precise values for Fig. width.This setup will avoid overfull boxes.
\caption{The explanatory case of noise within DRKG. \textbf{\textit{Chebi:28300}} and \textbf{\textit{DB00130}} indicate the same drug \textbf{\textit{L-Glutamine}}, but they are treated as different entities, which results in entity unalignment and facts missing. Meanwhile, the source document represents the gene \textbf{\textit{MSLN}} as a biomarker for cancer patients, and no confidence indicates it has a role in the disease \textbf{\textit{Mental Disorders}}, which introduces noisy interactions into the KG.}
\label{fig1}
\end{figure}

Previous methods utilized the topological properties of the integrated association networks (e.g., \textit{drug-disease-association} networks) to learn low-dimensional vector representations for predicting unknown interactions~\cite{luo2017network,wan2019neodti,saebi2022heterogeneous}. These methods adopted network-based models, which ignore the semantic relations between various entities (e.g., drug, pathway, disease). 
Subsequently, a line of works applied the knowledge graph embedding methods to learn the semantic relations with multi-dimensional embeddings for predicting DTI~\cite{mohamed2020discovering} and DDI~\cite{celebi2019evaluation,su2022biomedical}.
However, these approaches often struggle to learn the topological structure of complex biomedical knowledge graphs (KGs) efficiently.
\textcolor{black}{More recently, various models~\cite{lin2020kgnn,yu2021sumgnn,ma2022kg,liang2023knowledge} utilizing heterogeneous graph neural networks have achieved promising results}. 
These methods focus on learning the semantic knowledge and local structure of the neighboring relational paths, which enables them to capture the semantic relations and tractable pathways surrounding the predicted interaction.

Despite their effectiveness, existing KG-based models suffer from noisy interactions and relations within biomedical KGs~\cite{huang2022trustworthy,zhu2023dfmke}. The majority of biomedical knowledge graphs are generated from unstructured text and multi-source databases using natural language processing technology~\cite{pujara2017sparsity,drkg2020}, which can lead to the presence of \textcolor{black}{misaligned entities and counterfactual links} within the KGs. 
% Most biomedical knowledge graphs are constructed from unstructured text and multi-source databases by using the technology of natural language processing ~\cite{pujara2017sparsity,drkg2020}, which can result in noisy facts and factual errors in the KGs. 
For example, as shown in Fig.~\ref{fig1}, \textcolor{black}{the different entities \textbf{\textit{Chebi:28300}} and \textbf{\textit{DB00130}} within DRKG represent the same drug \textbf{\textit{L-Glutamine}}\footnote{https://go.drugbank.com/drugs/DB00130}, which is misaligned. The source document represents the gene \textbf{\textit{MSLN}} is a biomarker of cancer patients, and no evidence indicates it has a role in the disease \textbf{\textit{Mental Disorders}}, introducing noise into the DRKG~\cite{drkg2020}}. In addition, the different relations \textit{drug\_treats\_disease} and \textit{drug\_inhibits\_disease} represent the same meaning ``\textit{one drug can therapy one disease}'' in the contexts of Fig.~\ref{fig1}, which brings inconsistent semantics.
% \textbf{\textit{Chebi:28300}} and \textbf{\textit{DB00130}} are different entities in DRKG~\cite{drkg2020} but they indicate the same drug \textbf{\textit{L-Glutamine}}\footnote{https://go.drugbank.com/drugs/DB00130}, which results in facts missing (e.g., the fact of the drug \textit{DB00130} treats the disease \textit{Liver Injury}).
% In addition, the source text indicates the gene \textbf{\textit{MSLN}} as a predictor for cancer patients~\cite{okla2018assessment}, and there is no evidence supporting its role in the disease \textbf{\textit{Mental Disorders}}. This may introduce a fact error into DRKG. 

In this case, current KG-based methods are ineffective due to the noisy interactions and inconsistent semantics. 
Based on the above observations, we propose a novel knowledge-enhanced denoising network, called BioKDN. We design a structure reliability learning module for the local subgraph guided by downstream tasks, to contain reliable interactions. \textcolor{black}{Inspired by the successful application of smoothing techniques for image denosing~\cite{ma2018deep,guo2019smooth} by blurring noisy pixels, we develop a smooth semantic preservation module that blurs the similar relations to keep consistent semantics and ignore task-irrelevant edges. This reduces the negative impact of noisy interactions and inconsistent semantics existing in the KG}. 
To further focus on the knowledge-enhanced informative interactions, we maximize the mutual information between the representations of reliable structure and smoothed semantics. \textcolor{black}{BioKDN improves the AUC-ROC and Micro-Recall by 2.19\% and 2.76\% respectively on the DrugBank dataset in both DTI and DDI prediction tasks.}

In summary, the contributions of BioKDN include:
\begin{itemize}
    % \item This is the first work that studies the noisy interactions existing in the biomedical KGs and reduces the negative impact of noises in an end-to-end manner.
    \item We approach knowledge-enhanced molecular interaction prediction from a new perspective by adaptively reducing the negative impact of noisy interactions.
    % To the best of our knowledge, we first propose a denoising model on biomedical KGs to alleviate the negative impact of noisy interactions and conflicts for link prediction. 
    \item We innovatively propose learning the reliable structure and smoothing semantics by blurring similar relations, which reduces the negative influence of noisy interactions and maintains consistent semantics.
    \item We emphasize knowledge-enhanced reliable interactions by maximizing the mutual information between the learned structure and smoothed semantics to efficiently drop information irrelated to downstream tasks.
    % Extensive experiments of the link prediction on relations \textit{drug-target-interaction} (DTI) and \textit{drug-drug-interaction} (DDI) domenstrate DenoisedLP outperforms the state-of-the-art methods on four benchmark datasets. 
    % \item To the best of our knowledge, this is the first work that proposes a denoising method to learn the reliable interactions and smooth semantics by blurring the sparse relations of biomedical KGs, which alleviates the negative influence of noisy interactions and conflicts.
    % \item We emphasize reliable interactions by maximizing the mutual information between the learned subgraph structure and smoothed semantic relations to efficiently drop information irrelated to downstream tasks.
    \item Extensive experiments of the DTI and DDI prediction on benchmark datasets and contaminated KG demonstrate that BioKDN outperforms the state-of-the-art baselines.
\end{itemize}

\section{Related Work}
\subsection{KG-based Molecular Interaction Prediction}
% The availability of biomedical knowledge graphs (KGs) and heterogeneous data encourages researchers to develop various computational methods to predict unknown drug-target interactions (DTIs). 
% The success of link prediction based on knowledge graph embedding methods in social networks[], and recommendation [] have motivated researchers to develop computational models for drug discovery. 
Molecular interaction prediction is increasingly adopted in biomedical knowledge graphs to identify unknown biological relations and interactions between various molecular entities~\cite{kishan2021predicting}. The line of work mainly focuses on the completion of DTI and DDI relations on KGs.
TriModel~\cite{mohamed2020discovering} 
and KG-DDI~\cite{celebi2019evaluation} 
proposed novel knowledge graph embedding models to learn the informative global structure and semantic knowledge for completing the relations of DTI and DDI, respectively. To obtain rich neighborhood information and semantic relations of KG, KGNN~\cite{lin2020kgnn} proposed a graph neural network to learn the structural relations, which enhances the prediction of the DDI relations. Subsequently, KGE-NFM~\cite{ye2021unified} developed a unified knowledge graph embedding framework to predict missing DTI links by combining the knowledge graph and recommendation system. 
However, these methods only consider the structure of the biomedical KGs. Recent methods proposed various fusion models to integrate the features of molecular graphs and KG embeddings for enhancing DTI~\cite{ma2022kg} and DDI~\cite{chen2021muffin} prediction.
To focus on the local structure of the predicted entity pairs, SumGNN~\cite{yu2021sumgnn} designed a new method to efficiently emphasize the subgraph structure of the \textcolor{black}{biomedical} KG, which aids the drug interaction prediction. GraIL~\cite{teru2020inductive} and SNRI~\cite{xu2022subgraph} proposed to model the enclosing subgraph structure and neighboring relational paths around the target triple to effectively predict unknown links. \textcolor{black}{Besides, MINES~\cite{liang2024mines} further introduces intercommuncation mechanism and performs prediction on neighbor-enhanced subgraph, which also achieves promising performance.
}
To filter out irrelevant entities, AdaProp~\cite{zhang2023adaprop} designed an incremental sampling mechanism to preserve promising targets. 
However, the presence of noise such as entity misalignment, and false positive triples in the KGs greatly degrades the performance of these methods. To address the above limitations, we develop reliable structure learning and smooth semantic preservation modules to denoise unreasonable interactions and maintain consistent semantics.

\subsection{Denoising Methods on Graphs}
% Contrastive learning is a type of self-supervised learning that trains an encoder to learn representations based on the mutual information between different views~\cite{oord2018representation,velivckovic2018deep}. This approach has shown promising results for various downstream applications~\cite{qiu2020gcc,wei2022contrastive}. 
Denoising on graphs has been successfully applied to the recommendation~\cite{fan2023graph} and social networks~\cite{quan2023robust}. RGCF~\cite{tian2022learning} proposed a self-supervised robust graph collaborative filtering model to denoise unreliable interactions and preserve the diversity in a contrastive way for the recommendation. Similarly, SGDL~\cite{gao2022self} provided a universal solution using self-guided learning to denoise implicit noisy feedback that can be generalized to various recommendation tasks. 
However, these methods are limited in their ability to denoise noisy interactions with positive and negative feedback in domain-specific networks, and difficult to work on the complex biomedical KGs. To tackle these limitations, inspired by the smoothing insight in image denoising~\cite{ma2018deep,guo2019smooth}, we blur the complex relations to contain consistent semantics and learn the reliable interactions from the local subgraph.
% However, these methods only denoise noisy interactions with positive or negative feedback in the domain-specifical networks, and cannot consider the complex relationships in the biomedical KGs. To tackle these issues, we use the complex semantic relations in KGs to guide the reliability learning of graph structures and denoise unreliable interactions with conflict relations in a contrastive way. 

\section{Methodology}
\begin{figure}[t]
\centering
\includegraphics[width=1\columnwidth]{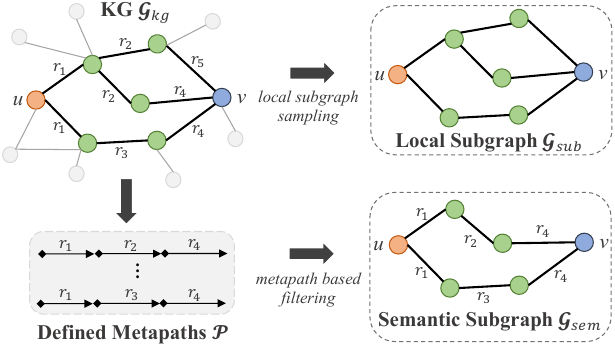} % Reduce the Fig. size so that it is slightly narrower than the column. Don't use precise values for Fig. width. This setup will avoid overfull boxes.
\caption{\textcolor{black}{An example of extracting local (structural reliability) and semantic subgraphs (semantic consistency) surrounding the target molecular pair $(u,v)$.}}
\label{fig1_example}
\end{figure}

% In this section, we elaborate on the proposed DenoisedLP method. We begin by introducing the preliminaries, followed by an overview of the framework. Finally, we discuss the training and learning strategy of the model in detail.
\begin{figure*}[t]
\centering
\includegraphics[width=0.95\textwidth]{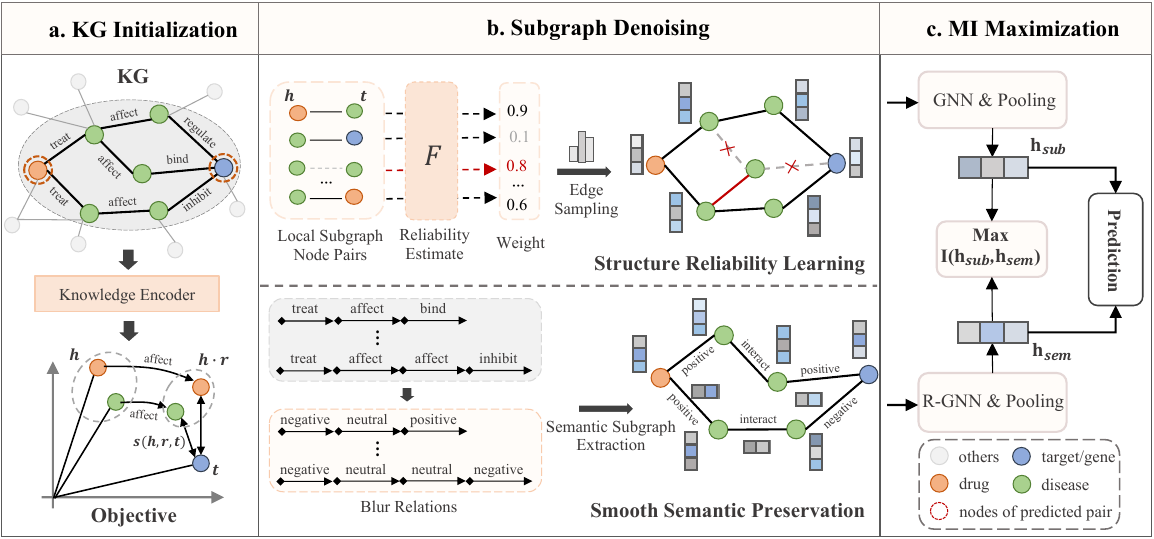} % Reduce the Fig. size so that it is slightly narrower than the column. Don't use precise values for Fig. width. This setup will avoid overfull boxes.
\caption{The BioKDN framework comprises three modules for predicting links in a given KG: (a) Initializing the entity and relation embeddings of the KG using RotatE; (b) Denoising the noisy interactions around the predicted link by learning reliable structure and preserving consistent semantics; (c) Maximizing the mutual information between refined and semantic subgraphs to focus on informative interactions.}
\label{fig2}
\end{figure*}
\subsection{Preliminaries}
\subsubsection{Biomedical Knowledge Graph} We define a biomedical knowledge graph DRKG~\cite{drkg2020} as $\mathcal{G}_{kg} = \{(h, r , t)|h,t\in \mathcal{E}, r\in \mathcal{R}\}$ where each triple $(h,r,t)$ describes a relation $r$ (e.g., DTI and DDI) between the biomedical entities $h$ and $t$. 
% with different node types $e_h$ and $e_t$ (e.g., drugs and genes) as a fact. 
% In addition, the concept of \textit{noise} in the KGs refers to the triples with conflicted relations and entities~\cite{ma2022ptruste,huang2022trustworthy}.

% This DRKG as a drug repurposing data source provides a large number of facts that can be used for downstream analysis tasks (e.g., drug-target interaction prediction and drug interaction identification).
% Note that the known triples with \textit{drug-target} and \textit{drug-drug} relationships are removed from the KG and stored separately as $P_{test} = \{(u,v)|u,v \in \mathcal{E}\}$ for the evaluation of downstream tasks.

\subsubsection{Local Subgraph}
Based on GraIL~\cite{teru2020inductive}, when given a KG $\mathcal{G}_{kg}$ and a molecular pair $(u,v)$, we extract a local subgraph surrounding the target link. Initially, we obtain the $k$-hop neighboring nodes $\mathcal{N}_{k}(u)=\{s|d(u,s)\leq k\}$ and $\mathcal{N}_{k}(v)=\{s|d(v,s)\leq k\}$ for both $u$ and $v$, where $d(\cdot,\cdot)$ represents the shortest path distance between target pair on $\mathcal{G}_{kg}$. We then obtain the set of nodes $V=\{s|s\in \mathcal{N}_{k}(u)\cap \mathcal{N}_{k}(v)\}$ as vertices of the local subgraph. Finally, we extract the edges $E$ linked by the set of nodes $V$ from $\mathcal{G}_{kg}$ as the local subgraph  $\mathcal{G}_{sub}=(V, E)$. We show an example in Fig.~\ref{fig1_example}.
% However, unlike GraIL, we treat the extracted subgraph as isomorphic and ignore the specific relations among nodes to focus on the link reliability of the subgraph.

\subsubsection{Semantic Subgraph}
% Given a molecular pair $(u,v)$ with the scheme $(e_u,r,e_v)$, we define a set of metapaths $\mathcal{P}=\{(r_{e_u},r_1,...,r_k,r_{e_v})|r_{e_u}\in \mathcal{N}^{rel}(e_u), r_k\in \mathcal{R}, r_{e_v} \in \mathcal{N}^{rel}(e_v)\}$ between nodes $u$ and $v$, where $\mathcal{N}^{rel}(\cdot)$ are the neighboring relations around target entity type and $k$ denotes the path length. According to the defined set of metapaths, we can extract various relational paths around the given link from original KG $\mathcal{G}_{kg}$ to construct a semantic subgraph $\mathcal{G}_{sem} = \{(h,r_{(h,t)},t)|h,t\in \mathcal{E},r_{(h,t)}\in \mathcal{R}_{(e_h, e_t)}\}$, where $\mathcal{R}_{(e_h, e_t)}$ denotes the relations between entity type $e_h$ and $e_t$.
\textcolor{black}{Given a molecular pair $(u,v)$, we define a set of metapaths $\mathcal{P}=\{P_1,\cdots,P_i,\cdots,P_n\}$ between nodes $u$ and $v$, where $n$ is the number of defined metapaths and $P_i=(\xrightarrow{r_1},\cdots,\xrightarrow{r_k})$. According to the defined set of metapaths, we can extract various relational paths around the given link from original KG $\mathcal{G}_{kg}$ to construct a semantic subgraph $\mathcal{G}_{sem}$ as shown in Fig.~\ref{fig1_example}.} 
% = \{(h,r_{(h,t)},t)|h,t\in \mathcal{E},r_{(h,t)}\in \mathcal{R}_{(e_h, e_t)}\}$, where $\mathcal{R}_{(e_h, e_t)}$ denotes the relations between entity type $e_h$ and $e_t$.

\subsubsection{Problem Definition} In this paper, we focus on predicting the missing molecular interactions based on biomedical KG $\mathcal{G}_{kg}$ by adaptively extracting high-quality facts and offering additional knowledge. We treat the molecular interaction prediction as a classification task, aiming to estimate the interaction probability of various relations (e.g., DTI and DDI). For a given unknown molecular pair $(u,v)$ with relation $r$, we propose a model to predict the interaction probability denoted as $p_{(u,r,v)} = \mathcal{F}((u,r,v)|\Theta, \mathcal{G}_{kg}, \mathcal{G}_{sub}, \mathcal{G}_{sem})$ by maximizing the mutual information between the local and semantic subgraphs, where $r$ represents the DTI or DDI.

% \subsection{Overview of DenoisedLP}
% The proposed DenoisedLP consists of three modules: the initialization of KG, the subgraph denoising module, and maximizing the mutual information between structure and semantic subgraphs, as illustrated in Fig.~\ref{fig2}. Given an initialized biomedical KG and a link to be predicted, we first extract the local and semantic subgraphs using their relevant local structures. 
% To denoise errors and relink reliable interactions, we propose structure reliability learning based on the local subgraph, resulting in an informative subgraph structure. Meanwhile, we construct the semantic subgraph to filter out task-independent triples and  preserve key semantic relations connected between the given link. To guide the structure learning of local subgraphs and preserve the semantic information, we maximize the mutual information between reliable structural and semantic knowledge.
% Finally, we use the learned representations of local and semantic subgraphs to predict the interaction probability of the given link.

\subsection{Initialization of Knowledge Graph}\label{sec:kge}
In this paper, we utilize DRKG~\cite{drkg2020} as our external biomedical KG. DRKG contains complex relationships between biological entities (e.g., symmetric and inverse interactions among genes). To effectively learn the semantic knowledge within the DRKG, we use the knowledge graph embedding method with relation rotation following in~\cite{sun2019rotate}. Given a triple $(h,r,t)$, we expect that $\mathbf{x}_t = \mathbf{x}_h\odot \mathbf{e}_r$, \textcolor{black}{where the $\odot$ represents the element-wise product. $\mathbf{x}_h,\mathbf{x}_t$ and $\mathbf{e}_r$ represent the embeddings of entities $h,r$ and the relation $r$, respectively.}
The score function is defined as follows:
\begin{equation}
    s(h,r,t) = || \mathbf{x}_h\odot \mathbf{e}_r - \mathbf{x}_t ||,
\end{equation}
% where the $\odot$ represents the element-wise product.
By minimizing the score of positive triples and maximizing the score of negative ones, we obtain the entity and relation embeddings $\mathbf{X}$ and $\mathbf{E}$ as the initial features of the KG.

\subsection{Subgraph Denoising}
% Most of the existing biomedical KGs are extracted from literature text (e.g., PubMed\footnote{https://pubmed.ncbi.nlm.nih.gov/}). Due to the limitation of the unstructured data and the ability of mining algorithms, KGs contain various noises, which decreases the performance of downstream link prediction tasks~\cite{pujara2017sparsity}. We propose a denoising model to learn reliable links and smooth noisy relations.

\subsubsection{Structure Reliability Learning}
To enable robust estimation of noisy interactions in KGs, we propose a structural reliability learning module for the local subgraph. This module can dynamically adjust the reliable subgraph structure by using the pre-trained node features and the feedback of the downstream prediction tasks.
% which can dynamically adjust the reliable subgraph structure by using the pre-trained node features and the feedback of the downstream prediction tasks. 
\textcolor{black}{The underlying assumption is that nodes with similar features or structures are more likely to interact with each other than those with irrelevant features or structures~\cite{zhang2020gnnguard,li2024gslb}}. 
Our objective is to assign weights to all edges between the set of nodes using a reliability estimation function denoted as $F(\cdot,\cdot)$, which relies on pre-trained node features.
% To achieve this, we aim to weighted all edges between the nodes set using reliability estimation function $F(\cdot,\cdot)$ based on the pretrained node features. 
Then, the refined local subgraph can be generated by filtering out noisy edges with low weight and retaining the reliable links with larger ones, as shown in Fig. \ref{fig2}b.
Specifically, given an extracted local subgraph $\mathcal{G}_{sub}=(V, E)$ around the molecular pair $(u, v)$, we model all possible edges between the nodes as a set of mutually independent Bernoulli random variables parameterized by the learned attention weights $\pi$.
\begin{equation}\label{eq2}
    \mathcal{G}_{sub}^{'}=\bigcup_{i,j\in V}\left\{(i,j)\sim\mathrm{Ber}\left(\pi_{i,j}\right)\right\}.
\end{equation}
Here, $V$ represents the set of nodes within the local subgraph and $(i,j)\in E$ denotes the edge between nodes $i$ and $j$. We optimize the reliability probability $\pi$ jointly with the downstream molecular interaction prediction tasks. The value of $\pi_{i,j}$ describes the task-specific reliability of edge $(i,j)$ where smaller $\pi_{i,j}$ indicates that the edge $(i,j)$ is more likely to be noised and should be assigned a lower weight or be removed. For each edge between node pair $(i,j)$, the reliable probability $\pi_{i,j} = F(i,j)$ can be calculated as follows:
\begin{equation}
    \begin{aligned}
    &\pi_{i,j} =\mathrm{sigmoid}\left(Z(i)Z(j)^\mathrm{T}\right),
    \\
    &Z(i) =\mathbf{MLP}\left(\mathbf{X}\left(i\right)\right),
\end{aligned}
\end{equation}
where $\mathbf{X}\left(i\right)$ represents the pre-trained feature of \textcolor{black}{node $i$ derived from Section~\ref{sec:kge}}, $Z(i)$ is the learned embedding of node feature $\mathbf{X}\left(i\right)$, and $\mathbf{MLP}\left(\cdot\right)$ denotes a two-layer perceptron in this work. Since the extracted local subgraph $\mathcal{G}_{sub}$ is not differentiable with the probability $\pi$ as Bernoulli distribution, \textcolor{black}{we use the reparameterization trick and relax the distribution~\cite{jang2016categorical,fan2023graph} $\mathrm{Ber}(\pi_{i,j})$ in Equation~(\ref{eq2}) as follows}:
\begin{equation}
    % \small
    \mathrm{Ber}(\pi_{i,j})\approx\mathrm{sigmoid}\left(\frac1t\left(\log\frac{\pi_{i,j}}{1-\pi_{i,j}}+\log\frac\epsilon{1-\epsilon}\right)\right),
\end{equation}
where $\epsilon\sim$ \textit{Uniform}$(0,1)$, $t\in \mathbb{R}^+$ indicates the temperature for the concrete distribution. \textcolor{black}{After relaxation, the binary entries $(i, j)$ sampled from a Bernoulli distribution are converted into a deterministic function of $\pi_{i,j}$ and $\epsilon$.}
With $t\textgreater0$, the function is smoothed with a well-defined gradient $\frac{\partial\mathrm{Ber}(\pi_{i,j})}{\partial \pi_{i,j}}$ that enables the optimization of learnable subgraph structure during the training process. The subgraph structure after the concrete relaxation is a weighted fully connected graph, which is computationally expensive. We hence drop the edges of the subgraph with a probability of less than 0.5 and get the refined graph $\mathcal{G}_{sub}^{'} = (V, E^{'})$. Subsequently, we perform the $L$-layer GCNs~\cite{kipf2016semi} on the refined subgraph with pre-trained node features to obtain its global representation $\mathbf{h}_{sub}$ as follows:
\begin{equation}
\begin{aligned}
        &h^l =\mathbf{GCN}\left(h^{l-1}, \mathcal{G}_{sub}^{'}\right),
         \\
        &\mathbf{h}_{sub}=\frac1{|V|}\sum_{i\in V}^{V}\sigma(f(h^L(i))),
\end{aligned}
\end{equation}
where the initial $h^0=\mathbf{X}$ and $\sigma(\cdot)$ represents the activation function. \textcolor{black}{$f(\cdot)$ denotes the feature transformation operation}.

% By using this method, we aim to learn local subgraph structures that capture the essential information for link prediction. This approach allows us to filter out noisy edges effectively and retain only the most informative ones.

\subsubsection{Smooth Semantic Preservation}
Biomedical KGs often contain noise and inconsistent relations (e.g., the different relations \textit{drug\_treats\_disease} and \textit{drug\_inhibits\_disease} represent the same semantic ``\textit{drug can therapy disease}'') that bring negative impact into the downstream molecular interaction prediction tasks~\cite{pujara2017sparsity}.
Inspired by the smoothing insight of image denoising~\cite{ma2018deep,guo2019smooth}, we design a knowledge-enhanced smooth semantic preservation module, which blurs similar relations and preserves the smoothed relational semantics to mitigate the adverse impact of potential inconsistency. Specifically, we utilize prior knowledge to generalize the interactions between biological entities into positive, neutral, and negative semantics based on the consistent meaning of different relations, thereby smoothing the KG. Subsequently, we develop a semantic subgraph extraction module to explore the neighboring relations by extracting the paths with pre-defined metapaths (i.e., patterns among drug, disease, and gene entities). 
\textcolor{black}{Given the molecular pair $(u, v)$, we use the defined metapaths to extract relational paths and construct a semantic subgraph $\mathcal{G}_{sem}$ as shown in Fig.~\ref{fig1_example}.} 
After obtaining the semantic subgraph, we design a $L$-layer relational graph neural network (R-GNN) inspired by~\cite{schlichtkrull2018modeling,xu2022subgraph} to obtain the global semantic representation of $\mathcal{G}_{sem}$. We define the updating function of the nodes in $l$-th layer as:
\begin{equation}
    \begin{gathered}\mathbf{x}_i^l=\sum_{r\in \mathcal{R}}\sum_{j\in\mathcal{N}_r(i)}\alpha_{i,r}\mathbf{W}_r^l\phi(\mathbf{e}_r^{l-1},\mathbf{x}_j^{l-1}),\\
    \alpha_{i,r}=\mathrm{sigmoid}\left(\mathbf{W}_1\left[\mathbf{x}_i^{l-1}\oplus\mathbf{x}_j^{l-1}\oplus\mathbf{e}_r^{l-1}\right]\right),
    \end{gathered}
\end{equation}
where $\mathcal{N}_r(i)$ and $\alpha_{i,r}$ denote the neighbors and the weight of node $i$ under the relation $r$, respectively. $\oplus$ indicates the concatenation operation. $\mathbf{W}_r^l$ represents the transformation matrix of relation $r$, $\phi$ is the aggregation operation $\phi(\mathbf{x},\mathbf{e}) = \mathbf{x}-\mathbf{e}$ to fuse the hidden features of nodes and relations, \textcolor{black}{and $\mathbf{e}_r^{l-1}$ is optimized from $\mathbf{e}_r^{0}$ layer by layer}. In addition, we initialize the node feature $\mathbf{x}_i^0$ and relation representation $\mathbf{e}_r^0$ using the pretrained embeddings $\mathbf{X}$ and $\mathbf{E}$. Finally, we obtain the global representation $\mathbf{h}_{sem}$ of the semantic subgraph $\mathcal{G}_{sem}$ as follows:
\begin{equation}
\mathbf{h}_{sem}=\frac1{|V_{sem}|}\sum_{i\in V_{sem}}^{V_{sem}}\sigma(f(\mathbf{x}_i^L)), 
\end{equation}
where $V_{sem}$ is the node set of semantic subgraph $\mathcal{G}_{sem}$. For more information about the smoothed KG and pre-defined metapaths, please refer to Section~\ref{metapaths}.

\subsection{MI Maximization in Subgraphs}
% To better guide the denoising process relies on the link prediction objective and balance the trust of structure reliability learning and the information of semantic graphs, we designed an auxiliary self-supervised task based on MIM to make the two kinds of graph representation
% To preserve the knowledge-enhanced reliable interactions, we 
% To cooperate in denoising noise contained in the KG from different views of the reliable structure and smoothed semantic relations, we design an auxiliary self-supervised task based on mutual information (MI) maximization. 
To preserve the knowledge-enhanced reliable interactions, we incorporate fusion information from diverse perspectives of the reliable structure and smoothed semantic relations through the maximization of mutual information (MI).
% We seek that the knowledge-enhanced relation smoothing blur task-irrelevant edges and cooperate with the structure reliability learning for denoising noisy interactions.
Specifically, we adopt InfoNCE~\cite{oord2018representation} to estimate mutual information between the representations of local structure and semantic subgraphs globally. 
In a formal context, when discussing the concept of subgraph mutual information, we treat the representations originating from both the reliable structure and the semantic subgraph, each extracted from a common link, as positive pairs. Conversely, the representations stemming from two distinct links within the refined local structure and the smoothed semantic are treated as adversarial pairs:
% Formally, for the subgraph mutual information, we consider the representations of the reliable structure and semantic subgraphs extracted from the same link as the positive pairs, while representations of two different links in the refined local structure and smoothed semantic as the antagonistic pairs:
\begin{equation}
% \small
% I(\mathbf{h}_{sub};\mathbf{h}_{sem})=-\sum_{m\in P}\log\frac{\exp(d(\mathbf{h}_{sub}^m,\mathbf{h}_{\footnotesize sem}^m))}{\sum_{n\in P}\exp(d(\mathbf{h}_{sub}^m,\mathbf{h}_{\footnotesize sem}^n))},
I(\mathbf{h}_{sub};\mathbf{h}_{sem})=-\log\frac{\exp(d(\mathbf{h}_{sub},\mathbf{h}_{\footnotesize sem})/\tau)}{\sum_{m\in P}\exp(d(\mathbf{h}_{sub},\mathbf{h}_{\footnotesize sem}^m)/\tau)},
\end{equation}
where $d(\cdot,\cdot)$ is set as a cosine similarity function to measure the similarity between two representation vectors and $\tau$ is a hyper-parameter indicating the temperature; $P$ represents all link pairs to be predicted and $\mathbf{h}_{sem}^m$ denotes the representation of global semantic subgraph for the link $m$.

\subsection{Prediction and Optimization}
We consider the molecular interaction prediction a classification task. For a given molecular pair $(u,v)$ with relation $r$, we model the interaction probability $p_{(u,r,v)}$ of the pair by adopting the learned representations from the structure and knowledge-enhanced semantic views as follows:
\begin{equation}
    p_{(u,r,v)} = \sigma(f([\mathbf{h}_{sub}\oplus\mathbf{h}_{sem}])),
\end{equation}
where the $\oplus$ indicates the concatenate operation. We then adopt the cross-entropy loss:
\begin{equation}
    % \ell(u,v)=-(y_{uv}\log p_{uv} + (1-y_{uv})\log (1-p_{uv})),
    \ell(u,v)=-\sum_{r\in \mathcal{R}}\log(p_{(u,r,v)})y_{(u,r,v)},
\end{equation}
where $y_{(u,r,v)}$ is the real label of the given link.
To denoise unreliable interactions from the structure and semantic subgraphs, we jointly optimize the link prediction task and the self-supervised MI maximization contrastive learning:
\begin{equation}
    \ell_\mathrm{total}(u,v)=\ell(u,v)+\lambda I(\mathbf{h}_{sub};\mathbf{h}_{sem}),
\end{equation}
where $\lambda$ is a hyper-parameter that weighs the contribution of the self-supervised MI mechanism. 
% The overall process is shown in Algorithm~\ref{alg:algorithm}.
% \begin{algorithm}[tb]
% \caption{DenoisedLP algorithm}
% \label{alg:algorithm}
% \textbf{Input}: Knowledge graph $G_{kg}$; Matrix $\mathbf{Y}$ of relation $r$\\
% \textbf{Parameter}: Optional list of parameters\\
% \textbf{Output}: Your algorithm's output
% \begin{algorithmic}[1] %[1] enables line numbers
% \STATE Let $t=0$.
% \WHILE{condition}
% \STATE Do some action.
% \IF {conditional}
% \STATE Perform task A.
% \ELSE
% \STATE Perform task B.
% \ENDIF
% \ENDWHILE
% \STATE \textbf{return} solution
% \end{algorithmic}
% \end{algorithm}

% \subsection{Theoretical Analysis}
% We can show that the ReliableMIP is theoretically effective in removing irrelevant facts. 

\subsection{Computational Complexity of BioKDN}
BioKDN consists of three main modules: structural reliability learning (SRL), smooth semantic preservation (SSP), and mutual information maximization (MI). The SRL contains a graph structure learning model whose computational complexity is $kD\cdot d+(kD)^2\cdot d$, where $k$ is the subgraph size, $D$ denotes the average degree of DRKG, $d$ is the embedding dimension, and $kD$ represents the average number of nodes within the subgraph. The SSP contains an RGNN model over subgraphs whose computational complexity is $kD+\frac{kD^2}{2}+(kD)\cdot d\cdot L$, where $L$ is the number of layers. The MI module contains an infoNCE loss computation whose computational complexity is $(N\cdot B+N)d$, where $N$ is the number of samples and $B$ denotes the batch size. The overall computational complexity of BioKDN is $kD((L+2)d^2+\frac{D}{2}+1)+(N\cdot B+N)d$. This shows the complexity of the BioKDN approximation polynomial, and its efficiency depends mainly on the sample size, the batch size, the average degree of KG, and the embedding dimension. Therefore, BioKDN can be scaled to large-scale sparse KGs (low average node degree), and the training efficiency can be balanced for large-scale datasets by reducing the embedding dimension and batch size.

\subsection{Discussion of Smoothing Operation}
Biomedical knowledge-enhanced molecular interaction prediction has several key applications such as drug-target interaction (DTI) prediction and drug-drug interaction (DDI) prediction, which is critical for drug discovery and clinical research. However, most biomedical KGs are constructed from publication text and multi-source databases that often contain unreliable interactions and inconsistent semantics of similar relations. \textcolor{black}{For example, the different relations \textit{drug\_treat\_disease} and \textit{drug\_inhibit\_disease} represent the same semantic ``\textit{a drug can therapy a disease}'' actually}. However, the KG embedding models will adopt different multi-dimensional vectors to represent their semantics. This will introduce the inconsistency of different semantic relations, which degrades the robustness of representation models.
Inspired by the smooth insights of image denoising~\cite{ma2018deep,guo2019smooth} by blurring noisy pixels, we smooth DRKG~\cite{drkg2020} by blurring inconsistent and sparse relations. The results in Section~\ref{comparison results} indicate this operation can effectively reduce the negative impact of inconsistent relations. 
\section{Experiments}
In this section, BioKDN\footnote{Code: \url{https://github.com/xiaomingaaa/BioKDN}} performs the molecular interaction prediction task for two key relations (i.e., \textit{drug-target interaction} and \textit{drug-drug interaction}) on biomedical KG.
\subsection{Experimental Setups}
\subsubsection{Datasets} 
For link prediction of the relation DTI, we empirically perform experiments on two real-world datasets: (1) \textit{\textbf{DrugBank}}~\cite{wishart2018drugbank} collects the unique bioinformatics and cheminformatics resources that contain 12,063 drug-target pairs with 2,515 drugs, and 2,972 targets. (2) \textit{\textbf{DrugCentral}}~\cite{avram2023drugcentral} is a drug database built from multiple sources, which contains 9,317 interactions between 1,061 drugs and 1,388 targets. For DDI prediction, we evaluate BioKDN on two wide-used datasets: (3) \textit{\textbf{DrugBank}}~\cite{wishart2018drugbank} contains 191,984 drug pairs with 86 types associated pharmacological relations for 1,703 drugs (e.g., \textit{increase of cardiotoxic activity}). (4) \textit{\textbf{TWOSIDES}}~\cite{tatonetti2012data} dataset contains 335 drugs with 26,443 drug pairs for 200 various side effect types. Following~\cite{zitnik2018modeling}, we ensure each DDI type has at least 900 drug pairs by keeping 200 commonly occurring types. We adopt the comprehensive DRKG~\cite{drkg2020} as the biomedical knowledge graph, which contains 97,238 entities and 5,874,261 triples. To smooth the semantic relations and filter out task-irrelevant edges of the DRKG, we blur the interactions between drugs, genes, and diseases into three types according to the semantic similarity of various relations. 

\begin{table}[t]
\centering
\caption{The blurred and original relations of DRKG.}
\label{tab:blurred_table}
\resizebox{\columnwidth}{!}{%
\begin{tabular}{l|l}
\toprule
\textbf{Blurred Relations}           & \textbf{Original Relations}\\ \midrule
Gene\_neutral\_Gene        & \begin{tabular}[c]{@{}l@{}}HumGenHumGen, VirGenHumGen, Q (production), E (affect)\\ Rg (regulation), I (signaling pathway), H (same protein)\end{tabular}                         \\ \midrule
Gene\_postive\_Gene         & V+ (activates), B (bind), E+ (increases expression), W (enhance)  \\ \midrule
Drug\_neutral\_Gene    & \begin{tabular}[c]{@{}l@{}}DrugVirGen, DrugHumGen,  E (affect), K (metabolism)\end{tabular}                              \\ \midrule
Drug\_positive\_Gene    & \begin{tabular}[c]{@{}l@{}}target, enzyme, carrier, A+ (agonism),\\ E+ (increase), B (bind), O (transport), Z (enzyme activity)\end{tabular}     \\ \midrule
Drug\_negative\_Gene    & N (inhibit), A- (antagonism), E- (decrease)                                   \\ \midrule
Gene\_negative\_Drug    & \begin{tabular}[c]{@{}l@{}}ANTAGONIST, INHIBITOR, BLOCKER, CHANNEL BLOCKER, \\ ANTIBODY\end{tabular}              \\ \midrule
Gene\_neutral\_Drug    & OTHER, ALLOSTERIC MODULATOR                  \\ \midrule
Gene\_positive\_Drug    & \begin{tabular}[c]{@{}l@{}}BINDER, MODULATOR, AGONIST, POSITIVE ALLOSTERIC \\ MODULATOR, ACTIVATOR, PARTIAL AGONIST\end{tabular} \\ \midrule

Drug\_positive\_Disease & \begin{tabular}[c]{@{}l@{}}Sa (side effect), J (role in disease pathogenesis)\end{tabular}                                        \\ \midrule
Drug\_negative\_Disease & \begin{tabular}[c]{@{}l@{}}treats, T (therapy), C (inhibits cell growth), \\Pa (alleviate), Pr (prevent)\end{tabular} \\ \midrule
Drug\_neutral\_Disease & Mp (biomarkers)                                          \\ \midrule
Gene\_positive\_Disease     & \begin{tabular}[c]{@{}l@{}}L (improper), U (causal), Y (polymorphisms), \\J (role in pathogenesis), G (promote), Ud (mutations)\end{tabular}                           \\ \midrule
Gene\_negative\_Disease     & Te (possible therapeutic effect)                                           \\ \midrule
Gene\_neutral\_Disease     & D (targets), Md (diagnostic), X (overexpression in disease)                                             \\ \midrule
Disease\_positive\_Gene     & DuG (upregulate)                                    \\ \midrule
Disease\_negative\_Gene     & DdG (downregulate)                                    \\ \midrule
Disease\_neutral\_Gene     & DaG (associate)                                         \\ \midrule
Disease\_neutral\_Disease  & DrD (resemble)  \\   \bottomrule                       
\end{tabular}%
}
\end{table}

\subsubsection{Evaluation Metrics}
In this paper, we consider the molecular interaction predictions as classification tasks (i.e., Binary classification for DTI prediction~\cite{ma2022kg} and Multi-class/type for DDI prediction~\cite{yu2021sumgnn}). To quantify the prediction performance of the two molecular interaction prediction tasks, we adopted the \textit{area under the receiver operating characteristic curve} (AUC-ROC), the \textit{area under the precision-recall curve} (AUC-PR) for binary and multi-type classification, and \textit{F1 score}, \textit{Recall score} for multi-class classification task. The details of these metrics are as follows:
\begin{itemize}
    \item \textbf{AUC-ROC} is calculated by $\sum_{k=1}^n TP_k\Delta FP_k$, where $k$ is $k$-th true-positive and false-positive operating point $TP_k, FP_k$, and $n$ represents the number of samples.
    \item \textbf{AUC-PR} is calculated by $\sum_{k=1}^n Prec_k\Delta Rec_k$, where $k$ is $k$-th true-positive and false-positive operating point $Prec_k, Rec_k$, $n$ indicates the number of samples.
    \item \textbf{F1 score}: average F1 score over different classes as $\frac{1}{N}\sum_{k=1}^N\frac{2P_k\cdot R_k}{P_k+R_k}$, where N is the number of classes and $P_k$, $R_k$ indicate the precision and recall for $k$-th class. This metric is more sensitive to the results for classes where samples are fewer. 
    \item \textbf{Recall score}: average Recall score over different classes as $\frac{1}{N}\sum_{k=1}^N\frac{TP_k}{TP_k + FN_k}$.
\end{itemize}
We perform 10-fold cross-validation on all datasets, and for each iteration, two blocks will be selected as the validation set and the test set respectively.
For binary and multi-type classification tasks, we select the best model based on the AUC-ROC of the validation set. In addition, we select the optimal model based on the F1 score of the validation set for the multi-class classification task.

\begin{table}[t]
\centering
\caption{The examples of pre-defined metapaths.}
\label{tab:my-table}
\resizebox{\columnwidth}{!}{%
\begin{tabular}{l|l}
\toprule
\textbf{Prediction Task} & \textbf{Metapaths}                         \\ \midrule
\begin{tabular}[c]{@{}l@{}}
drug-target interaction

\end{tabular}       & \begin{tabular}[c]{@{}l@{}}(drug\_neutral\_disease, disease\_neutral\_gene)\\ (drug\_neutral\_gene,   gene\_neutral\_gene),\\ (drug\_positive\_disease, disease\_neutral\_gene),\\ (drug\_negative\_disease, disease\_positive\_gene),\\
(drug\_negative\_gene, gene\_positive\_gene),\\
...
\end{tabular}    \\ \midrule
\begin{tabular}[c]{@{}l@{}}
drug-drug interaction
\end{tabular}
         & \begin{tabular}[c]{@{}l@{}}(drug\_neutral\_disease, disease\_neutral\_drug),\\  (drug\_neutral\_gene,   gene\_positive\_drug),\\  (drug\_positive\_disease, disease\_positive\_drug),\\
(drug\_negative\_disease, disease\_positive\_drug),\\
(drug\_negative\_gene, gene\_positive\_drug),\\
...
\end{tabular} \\ \bottomrule
\end{tabular}%
}
\end{table}

\subsubsection{Smoothed Relations}
\begin{figure}[t]
\centering
\includegraphics[width=0.9\columnwidth]{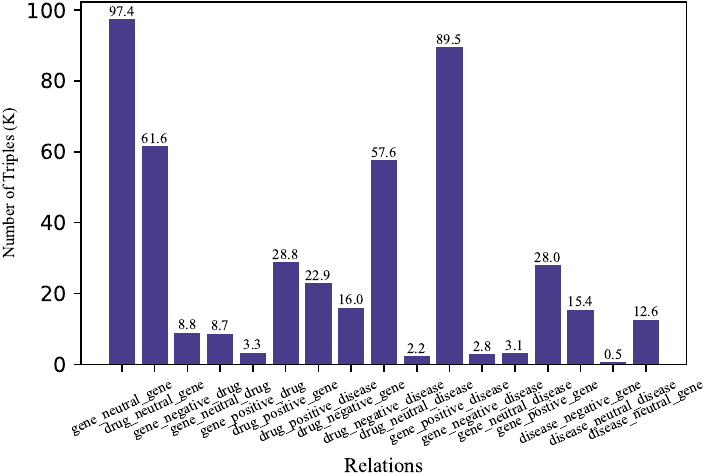}
\caption{The relation distribution of smoothed KG.}
\label{fig_ab}
\end{figure}

To effectively capture the consistent semantic information in DRKG and reduce the negative impact of noisy interactions and irrelevant edges on downstream tasks, we blur the relations according to the semantic similarity of the original relationship. As shown in Table~\ref{tab:blurred_table}, we filter out all entities except drug, gene (i.e., target), and disease, and classify the relationships between them into three categories (i.e., \textit{positive}, \textit{neutral}, and \textit{negative}). Take the original relations E+ and A+ as an example, they are similar as the semantic ``\textit{activation}'' and as the blurred association \textit{Drug\_positive\_Gene}, which represent drugs have a role in increasing the expression of genes. In Fig.~\ref{fig_ab}, the distribution of blurred relations of the smoothed KG is shown. We observe that the sparse links are blurred, which reduces the negative influence of noisy interactions.

\begin{table}[t]
\centering
\caption{Optimal hyper-parameter of BioKDN for each task.}
\label{tab:hyperparams}
\begin{tabular}{lll} \toprule
              & \textbf{DTI}   & \textbf{DDI}   \\ \midrule
learning rate & $10^{-2}$ & $10^{-2}$ \\
batch size    & 64    & 64    \\
embed dim     & 128   & 64    \\
$\lambda$             & 0.5   & 0.1   \\
$\pi$             & 0.5   & 0.3   \\
$\tau$             & 0.5   & 1   \\
$k$-hop             & 2   & 2   \\
KG initializer             & RotatE   & RotatE   \\ \bottomrule
\end{tabular}
\end{table}

\subsection{Data Preprocessing}
\subsubsection{Sample Details} For the DrugBank and DrugCentral datasets in predicting DTIs, we ensure the positive and negative samples for each drug are balanced by \textcolor{black}{generating them randomly}. Following SumGNN~\cite{yu2021sumgnn}, we keep the same settings by stratified sampling DDI from DrugBank for cross-validation. For the TWOSIDES dataset, we follow the method in~\cite{zitnik2018modeling} to generate negative counterparts for every positive sample from the unknown set of drug pairs. We perform 10-fold cross-validation and select the best model based on the AUC-ROC of the validation set. The average performance and standard deviation evaluated on the test set are reported. \textbf{To avoid data leakage}, we removed observable DTI and DDI facts from the external KG in advance.

\subsubsection{Entities Mapping}
DRKG provides the IDs for each entity that can link to widely used external databases (e.g., DrugBank IDs for drugs, Gene IDs for targets/proteins). Thus we link drugs and targets from the four datasets to corresponding entities in DRKG by using their IDs.

\subsubsection{Pre-defined Metapaths of DRKG}\label{metapaths}
To model relational paths and semantic information of DRKG efficiently, we define a set of reasonable metapaths for constructing semantic subgraphs. Specifically, we focus on the prediction of relations \textit{drug-target interaction} (DTI) and \textit{drug-drug interaction} (DDI). 
For DTI prediction, we extract semantic subgraphs using all $1$-hop metapaths between drug and target (i.e., gene). Similarly, in DDI prediction, we adopt $1$-hop metapaths between drug and drug as the relational schemes to extract semantic subgraphs. We illustrate some pre-defined metapaths in Table~\ref{tab:my-table}.
DRKG is a large-scale KG that integrates other widely-used biomedical KGs (e.g., Hetionet and GNBR). Thus DRKG contains 5,874,261 triples with 107 relations, which covers the relations from other KGs. We have categorized the biomedical relations from DRKG into \textit{positive}, \textit{neutral}, and \textit{negative} in Table~\ref{tab:blurred_table}. Thus we can map the relations of other biomedical KGs (e.g., Hetionet) directly using this mapping table, without re-manual processing.

\begin{table}[t]
\centering
\caption{The performance on DrugBank and DrugCentral for \textit{DTI} prediction. The \textbf{boldface} denotes the highest score and \underline{underline} indicates the second highest score.}
\label{tab:dti}
\resizebox{\columnwidth}{!}{%
\begin{tabular}{@{}ccccc@{}}
\toprule
\multicolumn{1}{c}{\multirow{2}{*}{Methods}} & \multicolumn{2}{c}{DrugBank} & \multicolumn{2}{c}{DrugCentral} \\ \cmidrule(l){2-5} 
\multicolumn{1}{c}{}                  & AUC-ROC     & AUC-PR     & AUC-ROC       & AUC-PR       \\ \midrule
GCN-KG       & 80.41 $\pm$ \text{\small 0.15} & 79.33 $\pm$ \text{\small 0.21} & 84.66 $\pm$ \text{\small 0.32} & 83.88 $\pm$ \text{\small 0.15} \\
% GraphDTA     & 81.62 $\pm$ \text{\small 0.12} & 81.92 $\pm$ \text{\small 0.61} & 85.34 $\pm$ \text{\small 0.11} & \underline{85.24 $\pm$ \text{\small 0.54}} \\ 
RotatE       & 77.65 $\pm$ \text{\small 0.31} & 75.99 $\pm$ \text{\small 0.13} & 81.18 $\pm$ \text{\small 0.62} & 81.03 $\pm$ \text{\small 0.11} \\
GraIL        & 80.54 $\pm$ \text{\small 0.17} & 81.37 $\pm$ \text{\small 0.37} &  82.74 $\pm$ \text{\small 0.45}        &  82.89 $\pm$ \text{\small 0.57} \\
TriModel     & 81.23 $\pm$ \text{\small 0.13} & 81.85 $\pm$ \text{\small 0.22} & 80.91 $\pm$ \text{\small 0.21} & 81.59 $\pm$ \text{\small 0.55} \\
SNRI         & 81.33 $\pm$ \text{\small 0.39} & 82.09 $\pm$ \text{\small 0.33} &  82.74 $\pm$ \text{\small 0.45}        &  82.89 $\pm$ \text{\small 0.57} \\
AdaProp      & 81.45 $\pm$ \text{\small 0.15} & 82.01 $\pm$ \text{\small 0.24} &  83.65 $\pm$ \text{\small 0.22}        &  83.27 $\pm$ \text{\small 0.36} \\
KG-MTL      & 82.55 $\pm$ \text{\small 0.31} & 81.79 $\pm$ \text{\small 0.52} & 84.39 $\pm$ \text{\small 0.55} & 83.13 $\pm$ \text{\small 0.74} \\
KGE-NFM     & \underline{82.71 $\pm$ \text{\small 0.22}} & \underline{82.09 $\pm$ \text{\small 0.52}}  & \underline{86.34 $\pm$ \text{\small 0.16}} & \underline{84.65 $\pm$ \text{\small 0.14}} \\ \midrule
BioKDN  & \textbf{84.90 $\pm$ \text{\small 0.35}} & \textbf{84.52 $\pm$ \text{\small 0.44}} & \textbf{88.79 $\pm$ \text{\small 0.23}} & \textbf{87.36 $\pm$ \text{\small 0.19}} \\

ours w/o SRL & 82.32 $\pm$ \text{\small 0.13}         & 82.55 $\pm$ \text{\small 0.11}          & 86.26 $\pm$ \text{\small 0.07}             & 85.24 $\pm$ \text{\small 0.12}             \\
ours w/o SSP & 81.78 $\pm$ \text{\small 0.23}         & 83.19 $\pm$ \text{\small 0.21}          & 86.54 $\pm$ \text{\small 0.14}             & 86.28 $\pm$ \text{\small 0.15}             \\
ours w/o MI & 82.05 $\pm$ \text{\small 0.09}         & 82.18 $\pm$ \text{\small 0.18}          & 86.34 $\pm$ \text{\small 0.09}             & 86.98 $\pm$ \text{\small 0.17}             \\ \bottomrule
\end{tabular}%
}
\end{table}

\subsection{Implementation Details of BioKDN} 
All the experiments in this work were conducted on a Linux server with an Intel Xeon(R) Platinum 8255C Processor (12 vCPU @2.5GHz), 43GB of RAM, and 2 RTX 2080Ti cards (11GB of RAM each). We implement BioKDN in Pytorch with Python 3.8.5. 
% Source code is available at \url{https://github.com/xiaomingaaa/BioKDN}.

To avoid data leakage, we removed observable DTI and DDI facts from the KG in advance.
% and they are invisible to BioKDN during training.
\textcolor{black}{The average performance and standard deviation evaluated on the test set are reported on Table~\ref{tab:dti} and Table~\ref{tab:ddi}}. We adopt the Xavier initializer~\cite{glorot2010understanding} to initialize the model parameters and optimize BioKDN with Adam~\cite{kingma2014adam}. The grid search is applied to retrieve optimal hyper-parameters. We tune the learning rate among $\{10^{-4},10^{-3},10^{-2},10^{-1}\}$, the embedding size in $\{32,64,128,256,512\}$, the size of subgraphs in $\{1, 2, 3, 4\}$-hop and the pruning threshold $\pi$ in $\{0.1,0.3,0.5,0.7,0.9\}$. Besides, we set weight $\lambda=0.5$ of the loss function. In addition, we search the weight $\lambda$ of the contrastive loss among $\{0.1, 0.3, 0.5, 0.7, 0.9, 1\}$ and the temperature $\tau$ among $\{0.5,1,2\}$. The final used hyperparameters of BioKDN for each prediction task are shown in Table~\ref{tab:hyperparams}.

\begin{table}[t]
\centering
\caption{The results comparison on DrugBank and TWOSIDES for DDI prediction. The best is marked with \textbf{boldface} and the second best is with \underline{underline}.}
\label{tab:ddi}
\resizebox{\columnwidth}{!}{%
\begin{tabular}{@{}ccccc@{}}
\toprule
\multicolumn{1}{c}{\multirow{2}{*}{Methods}} & \multicolumn{2}{c}{DrugBank (Multi-class)} & \multicolumn{2}{c}{TWOSIDES (Multi-label)} \\ \cmidrule(l){2-5} 
\multicolumn{1}{c}{}                  & Micro-F1     & Micro-Rec     & AUC-ROC       & AUC-PR       \\ \midrule
GCN-KG                              & 79.34 $\pm$ \text{\small 0.16}        & 82.56 $\pm$ \text{\small 0.23}          & 85.22 $\pm$ \text{\small 0.32}          & 82.57 $\pm$ \text{\small 0.12}         \\
RotatE                              & 76.41 $\pm$ \text{\small 0.11}        & 80.71 $\pm$ \text{\small 0.15}          & 85.92 $\pm$ \text{\small 0.18}          & 82.69 $\pm$ \text{\small 0.21}         \\
GraIL & 83.39 $\pm$ \text{\small 0.35}         & 76.11 $\pm$ \text{\small 0.46}          & 83.72 $\pm$ \text{\small 0.18}          & 80.73 $\pm$ \text{\small 0.09}             \\
KGNN                                  & 76.13 $\pm$ \text{\small 0.32}         & 74.62 $\pm$ \text{\small 0.42}          & 86.97 $\pm$ \text{\small 0.23}          & \underline{82.71 $\pm$ \text{\small 0.41}}        \\
SNRI           & 84.57 $\pm$ \text{\small 0.13}       & 82.13 $\pm$ \text{\small 0.19}          & 85.24 $\pm$ \text{\small 0.54}          & 81.75 $\pm$ \text{\small 0.27}         \\
AdaProp        & 83.26 $\pm$ \text{\small 0.10}       & 82.01 $\pm$ \text{\small 0.12}          & 86.41 $\pm$ \text{\small 0.13}          & 82.53 $\pm$ \text{\small 0.21}         \\
SumGNN                                & \underline{85.58 $\pm$ \text{\small 0.10} }        & \underline{82.79 $\pm$ \text{\small 0.19}}          & \underline{87.42 $\pm$ \text{\small 0.16}}          & 82.65 $\pm$ \text{\small 0.07}         \\ \midrule
BioKDN & \textbf{87.51} $\pm$ \text{\small 0.11}         & \textbf{85.55} $\pm$ \text{\small 0.13}          & \textbf{88.62} $\pm$ \text{\small 0.09}             & \textbf{84.73} $\pm$ \text{\small 0.12}             \\ 
ours w/o SRL & 85.67 $\pm$ \text{\small 0.19}         & 82.31 $\pm$ \text{\small 0.13}          & 86.21 $\pm$ \text{\small 0.21}             & 82.36 $\pm$ \text{\small 0.25}             \\
ours w/o SSP & 84.59 $\pm$ \text{\small 0.15}         & 83.01 $\pm$ \text{\small 0.19}          & 85.38 $\pm$ \text{\small 0.06}             & 83.12 $\pm$ \text{\small 0.22}             \\
ours w/o MI & 85.07 $\pm$ \text{\small 0.31}         & 82.87 $\pm$ \text{\small 0.24}          & 85.96 $\pm$ \text{\small 0.05}             & 82.81 $\pm$ \text{\small 0.32}             \\ \bottomrule
\end{tabular}%
}
\end{table}

\subsection{Baselines}
To verify the performance of BioKDN, we compare it against various baselines as follows:
\begin{itemize}
    % \item \textbf{LR}, \textbf{RF}, and \textbf{MLP} applied the molecular fingerprints (ECFP) of drugs and the PSC features of the target sequence. And MLP adopts a 3-layer perceptron with a hidden size of 256~\cite{10.1093/bioinformatics/btaa880}.
    \item \textbf{GCN-KG} and \textbf{RotatE}~\cite{sun2019rotate} adopted the graph neural network~\cite{kipf2016semi} and relational rotation in complex space to learn the embedding of entities and relations from the DRKG,  
    and then predicted the links for DTI or DDI using the embeddings.
    % \item \textbf{GraphDTA}~\cite{10.1093/bioinformatics/btaa921} applied graph neural network to encode molecule and used CNN to obtain the sequence features of the target.
    % \item  is a knowledge graph embedding (KGE) method by translating embeddings that models the semantic relations of entities including drugs and targets. 
    \item \textbf{GraIL}~\cite{teru2020inductive} utilized a local subgraph for inductive relation prediction on KGs. To model neighboring relations effectively, \textbf{SNRI}~\cite{xu2022subgraph} adopted the semantic subgraphs by extracting semantic relational paths to learn informative embedding. To filter out irrelevant entities, \textbf{AdaProp}~\cite{zhang2023adaprop} designed an incremental sampling mechanism to preserve the nearby targets.
    \item \textbf{TriModel}~\cite{mohamed2020discovering} and \textbf{KGE-NFM}~\cite{ye2021unified} developed new methods to learn the relational representation of entities and relations, then predict the unknown links for DTI. Additionally, \textbf{KG-MTL}~\cite{ma2022kg} proposed a global knowledge enhanced multi-task method to predict unknown DTI.
    % \item \textbf{TriModel}~\cite{mohamed2020discovering} and \textbf{KGE-NFM}~\cite{ye2021unified} developed new KGE methods to learn the representation of entities and relations, then predict the unknown links.
    \item \textbf{KGNN}~\cite{lin2020kgnn} aggregated neighborhood information for each node from their local receptive via GNN on the biomedical knowledge graph for link prediction of relation DDI. \textbf{SumGNN}~\cite{yu2021sumgnn} focused on extracting information from the local subgraph in a learnable way and considered the DDI prediction as multi-type and multi-class classification tasks.
\end{itemize}
Our approach focuses on subgraph-based reasoning on the biomedical knowledge graph to robustly predict molecular interactions. From the technical perspective, we compare the SOTA models based on knowledge graph representation (e.g., RotatE, 2018) and subgraph-based robust GNN (e.g., AdaProp, 2023 and SNRI, 2022). Meanwhile, we compare the optimal models (e.g., SumGNN, 2021, KGE-NFM, 2021, and KG-MTL, 2022) for KG-based molecular interaction prediction from the application perspective. This study provides a detailed and systematic comparison of notable methods in subgraph reasoning alongside KG-enhanced molecular interaction prediction models.

\subsection{Implementation Details of Baselines}
For baselines, we adopt official code packages from the authors for RotatE\footnote{github.com/DeepGraphLearning/KnowledgeGraphEmbedding}~\cite{sun2019rotate}, GraIL\footnote{https://github.com/kkteru/grail}~\cite{teru2020inductive}, SNRI\footnote{https://github.com/Tebmer/SNRI}~\cite{xu2022subgraph}, AdaProp\footnote{https://github.com/LARS-research/AdaProp}~\cite{zhang2023adaprop}, TriModel\footnote{The official source is not accessed, we requested the code from the author at: https://github.com/xiaomingaaa/BioKDN/baseline/trimodel.py}~\cite{mohamed2020discovering}, KGE-NFM\footnote{https://zenodo.org/records/5500305}~\cite{ye2021unified}, KGNN\footnote{https://github.com/xzenglab/KGNN}~\cite{lin2020kgnn} and SumGNN\footnote{https://github.com/yueyu1030/SumGNN}~\cite{yu2021sumgnn}. We use the GCN module of the DGL\footnote{https://github.com/dmlc/dgl} library on external DRKG~\cite{drkg2020} to implement GCN-KG and capture the embeddings of all molecules. For fair comparisons, we tune them for optimal performance.

% \noindent\textbf{Evaluation Metrics.} 
% For the DrugBank and DrugCentral datasets in predicting DTI, we ensure the positive and negative samples for each drug are balanced by random generating. Following SumGNN~\cite{yu2021sumgnn}, we keep the same settings by stratified sampling DDI from DrugBank for cross-validation. For the TWOSIDES dataset, we follow the method in~\cite{zitnik2018modeling} to generate negative counterparts for every positive sample from the unknown set of drug pairs. We perform 10-fold cross-validation and select the best model based on the AUC-ROC of the validation set. \textcolor{black}{The average performance and standard deviation evaluated on the test set are reported on Table~\ref{tab:dti} and Table~\ref{tab:ddi}}.

% we reproduce them on the \textbf{smoothed KG}.

\begin{figure*}[t]
\centering
\includegraphics[width=1\textwidth]{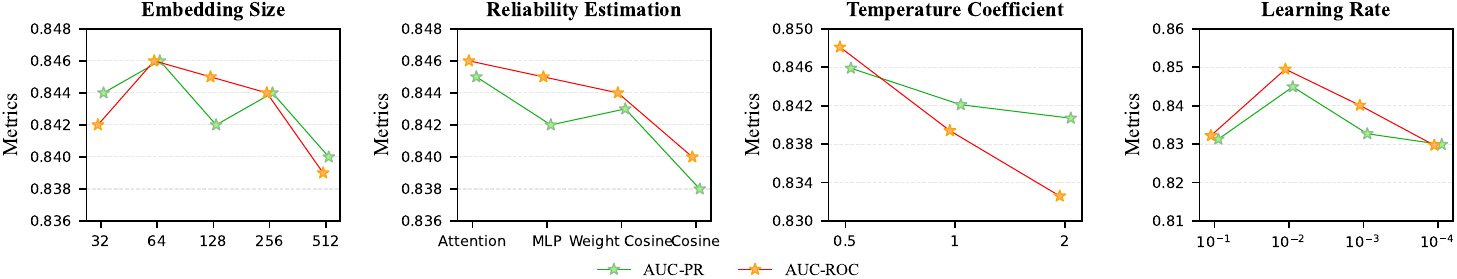} % Reduce 
% }
% \captionsetup[subFig.]{labelformat=empty}
\caption{Hyper-parameter sensitivity analysis of DTI prediction based on the DrugBank dataset.}

\label{fig_param_anay}
\end{figure*}

\subsection{Comparison with Baselines}\label{comparison results}
We report the performance of our model and baselines for predicting molecular interactions of the relations DTI and DDI in Table~\ref{tab:dti} and Table~\ref{tab:ddi}, respectively. 
As shown in Table~\ref{tab:dti}, we observe BioKDN achieves the best prediction results in the DTI task on both DrugBank and DrugCentral datasets. Specifically, BioKDN improves the AUC-ROC and AUC-PR by at least 2.19\% and 2.43\% respectively on the DrugBank dataset and achieves the 2.45\% and 2.71\% absolute increase over the best baseline on DrugCentral data. 
For the prediction of DDI, we find that the boosts of BioKDN on DrugBank for the multi-class task in Micro-F1 and Micro-Recall score up to 1.93\% and 2.76\% respectively. The performance of BioKDN on the TWOSIDES dataset has achieved 1.2\%, and 2.02\% improvement in AUC-ROC and AUC-PR compared with the best baseline.

Furthermore, we have the following observations:
(1) Compared with RotatE, the GCN-KG utilizing the neighboring information and structures achieves better performance on DTI and DDI prediction, which indicates that the neighboring structure benefits the downstream prediction tasks. (2) Compared with TriModel and KGNN, the SNRI using local semantic relations performs better than them on DrugBank for predicting DTI and DDI, which implies that the local semantic relations are more effective than the global structure and relations in predicting unknown molecular interactions. (3) Among the subgraph-based methods (i.e., GraIL, SNRI, and SumGNN), BioKDN can achieve superior improvement. This is because the noisy interactions in the local subgraph may make it hard for models to learn reliable neighborhood information effectively, degrading their performance. (4) 
BioKDN outperforms AdaProp in pruning task-irrelevant facts, indicating that knowledge-enhanced smoothing semantics can effectively improve the performance of denoising noisy interactions.
In short, the BioKDN reduces the negative influence of noise by denoising unreliable interactions in a learnable way and preserving knowledge-enhanced smoothing semantics, resulting in superior results.

\begin{table}[t] 
\centering
\caption{The performance of BioKDN on DrugBank for DTI prediction with various KGE methods.}
\label{kges}
\begin{tabular}{llll}
\toprule
Metrics & \textbf{TransE} & \textbf{\textcolor{black}{RotatE}} & \textbf{DistMult} \\ \midrule
AUC-ROC & 84.01           & \textbf{84.88}           & 84.23             \\
AUC-PR  & 83.87           & \textbf{84.37}           & 84.11       \\ \bottomrule     
\end{tabular}
\end{table}

\subsection{Ablation Study}
To investigate the impact of each module in BioKDN, we perform an ablation study on all datasets for DTI and DDI prediction by removing: (i) structure reliability learning (called \textbf{ours w/o SRL}), (ii) smooth semantic preservation (called \textbf{ours w/o SSP}), (iii) Mutual information (MI) maximization of dual-view subgraphs (called \textbf{ours w/o MI}), respectively. We can observe that all variants of BioKDN perform worse than the original model in Table~\ref{tab:dti} and Table~\ref{tab:ddi}, which verifies the effectiveness of each component.

\noindent\textbf{\underline{ours w/o SRL}.} We observe a significant reduction in performance across all datasets for DTI and DDI prediction after removing the structure reliability learning module. This suggests that the unreliable subgraph structure is less expressive for downstream tasks and cannot effectively eliminate the negative influence of noisy interactions. In contrast, a complete reliable structure can improve the performance of the original model by removing possible noise and retaining trustworthy interactions. 

\noindent\textbf{\underline{ours w/o SSP}.} From the results reported in Table~\ref{tab:dti} and Table~\ref{tab:ddi}, we notice a significant degradation in performance on all datasets when omitting the semantic subgraph of the smooth semantic preservation module. This observation demonstrates the effectiveness of the semantic subgraph, extracted by pre-defined metapaths, in keeping the consistent semantics of task-relevant relations. Intuitively, \textbf{ours w/o SRL} together with \textbf{ours w/o SSP} demonstrate the effectiveness of denoising unreliable interaction from the local structure and smooth semantic views.  

\noindent\textbf{\underline{ours w/o MI}.} Additionally, removing the MI maximization from BioKDN results in a reduction of performance on all datasets for both tasks. The results demonstrate that the learned reliable substructure, guided by knowledge-enhanced smoothing semantics and the removal of task-irrelevant relations, is effective in enhancing the performance of KG-based methods. These findings of the various variants show that the original model BioKDN can effectively enhance the superior performance of the link prediction tasks.

% \begin{Fig.}[t]
% \centering
% \includegraphics[width=0.9\columnwidth]{figs/ablation_study.pdf}
% \caption{Performance comparison of DTI and DDI predictions over different variants of DenoisedLP.}
% \label{fig_ab}
% \end{Fig.}

\subsection{Performance on Different KG initializers}
We used knowledge graph embedding (KGE) methods to initialize the vector representation of entities and relations. To find the best KGE methods, we performed BioKDN on TransE~\cite{transe}, RotatE~\cite{sun2019rotate}, and DistMult~\cite{distmult} for the DTI prediction task. The performance of them is reported in Table~\ref{kges}. We can observe that \textcolor{black}{RotatE} achieves the best result, which shows that modeling complex relations in advance is beneficial to prediction performance. Therefore, we adopt RotatE\footnote{We have provided details in Section~\ref{sec:kge}} to learn the vector representation of KG elements (e.g., drug/target molecules).

\begin{figure}[t]
\centering
\includegraphics[width=1\columnwidth]{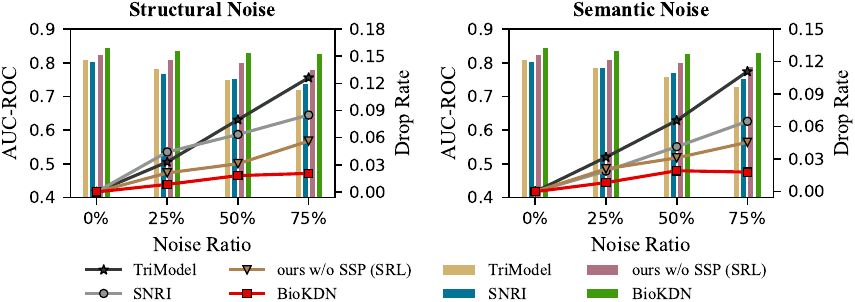} % Reduce the Fig. size so that it is slightly narrower than the column. Don't use precise values for Fig. width.This setup will avoid overfull boxes.
\caption{Performance over various noisy types with different ratios for DTI prediction. The bar represents the AUC-ROC and the line indicates the degradation ratio on the value. The \textbf{smaller} the drop rate, the better.}
\label{fig_noise}
\end{figure}

\begin{figure}[t]
\centering
\includegraphics[width=1\columnwidth]{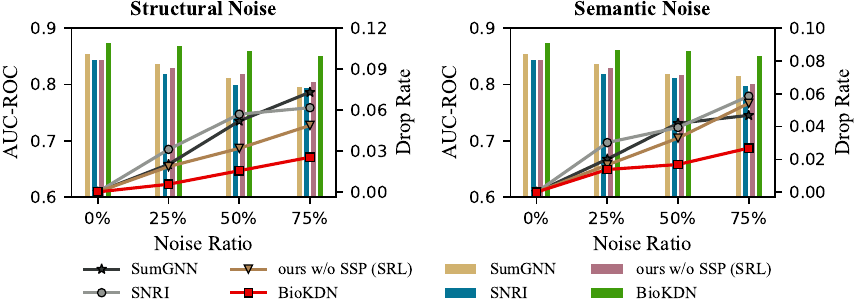} % Reduce the Fig. size so that it is slightly narrower than the column. Don't use precise values for Fig. width.This setup will avoid overfull boxes.
\caption{Performance over various noisy types with different ratios for DDI prediction. The bar represents the Micro-F1 and the line indicates the degradation ratio on the value. The \textbf{smaller} the drop rate, the better.}
\label{fig_noise_ddi}
\end{figure}
\subsection{Hyper-parameter Sensitivity}
% We conduct hyper-parameter analysis on Drugbank for DTI prediction. More details are shown in the Appendix.
% to study the influence of several hyper-parameters.
% Specifically, we choose the embedding size and the reliability estimation functions to analyze the sensitivity impact of them. 

\subsubsection{Impact of embedding size}
We explore the effect of hidden embedding size by varying it from 32 to 516. The left of Fig.~\ref{fig_param_anay} depicts the changing trend of the AUC-ROC and AUC-PR values on the DrugBank dataset evaluated on BioKDN. Based on the results, we observe that the AUC-ROC values of BioKDN variation across different embedding sizes collapsed into a hunchback shape. The reason could be that enough embedding size can represent more information, while the larger one will progressively introduce a lot of noise with the degradation of BioKDN. 

\subsubsection{Impact of reliability estimation}
To investigate the impact of various reliability estimation functions $F(\cdot,\cdot)$ defined in the Section \textbf{Structure Reliability Learning}, we conduct experiments by varying the estimation types to the \textit{Attention}, \textit{MLP}, \textit{Weighted\_Cosine}, and \textit{Cosine}. 
\textcolor{black}{As illustrated in Fig.~\ref{fig_param_anay}}, the \textit{Attention} with linear attention modeling the reliability weight between nodes set has the best performance. The \textit{MLP} adopts a 2-layer preceptor, which achieves a secondary best result by learning the weight of the node pairs from the extracted local subgraph. This is because the attention mechanism can better model the importance of node pairs compared with the \textit{MLP}, which improves the effectiveness of estimating the reliable edges.
Additionally, the parametric \textit{Weighted\_Cosine} is better than the non-parametric \textit{Cosine} indicating the learned weight guided by downstream tasks is more efficient.

\subsubsection{Impact of temperature coefficient $\tau$}
We evaluate the influence of different temperatures $\tau$ for BioKDN in Fig.~\ref{fig_param_anay}. We can see that as $\tau$ becomes larger, the effect of BioKDN gradually decreases. This indicates that a large $\tau$ will cause BioKDN to be unstable, thus making the effect worse. Therefore, we set $\tau=0.5$ for the DTI task.

\subsubsection{\textcolor{black}{Impact of learning rate}}
We conduct experiments to study the influence of the learning rate by varying it to be $10^{-1}$, $10^{-2}$, $10^{-3}$, and $10^{-4}$. The results show a hunchback shape in Fig.~\ref{fig_param_anay} and we observe that BioKDN achieves the best result when $10^{-2}$ is adopted. Finally, we set the learning rate as $10^{-2}$ for downstream tasks.

\subsubsection{\textcolor{black}{Impact of $\lambda$ and $\pi$}}
We conduct experiments to study the impact of the pruning threshold $\pi$ and the loss weight $\lambda$ of MI module shown in Table~\ref{tab:param}. $\lambda$ is tested on various scales when $\pi=0.1$ and $\pi$ is evaluated on various values when $\lambda=0.5$.
We can observe that the AUC-ROC values of BioKDN across different $\lambda$ and $\pi$ collapsed into a hunchback shape, which shows a balanced coordination is beneficial. Thus the performance of BioKDN was reported under $\lambda=0.5$ and $\pi=0.5$.

\begin{table}[t]
\centering
\caption{The AUC-ROC values of BioKDN on DrugBank for DTI prediction across different $\lambda$ and $\pi$.}
\label{tab:param}
\resizebox{\columnwidth}{!}{%
\begin{tabular}{ccccccc}
\toprule
\textbf{Parameter} & \textbf{0.1} & \textbf{0.3} & \textcolor{black}{\textbf{0.5}} & \textbf{0.7} & \textbf{0.9} & \textbf{1} \\ \midrule
\textbf{$\mathbf{\lambda}$}         & 83.11      & 83.97          & \textbf{84.78}          & 84.26          & 84.16          & 83.96        \\
\textbf{$\pi$}       & 84.13    & 84.01          & \textbf{84.83}          & 83.67          & 82.21          & -       \\ \bottomrule
\end{tabular}%
}
\end{table}

\begin{figure*}[t]
\centering
\includegraphics[width=1\textwidth]{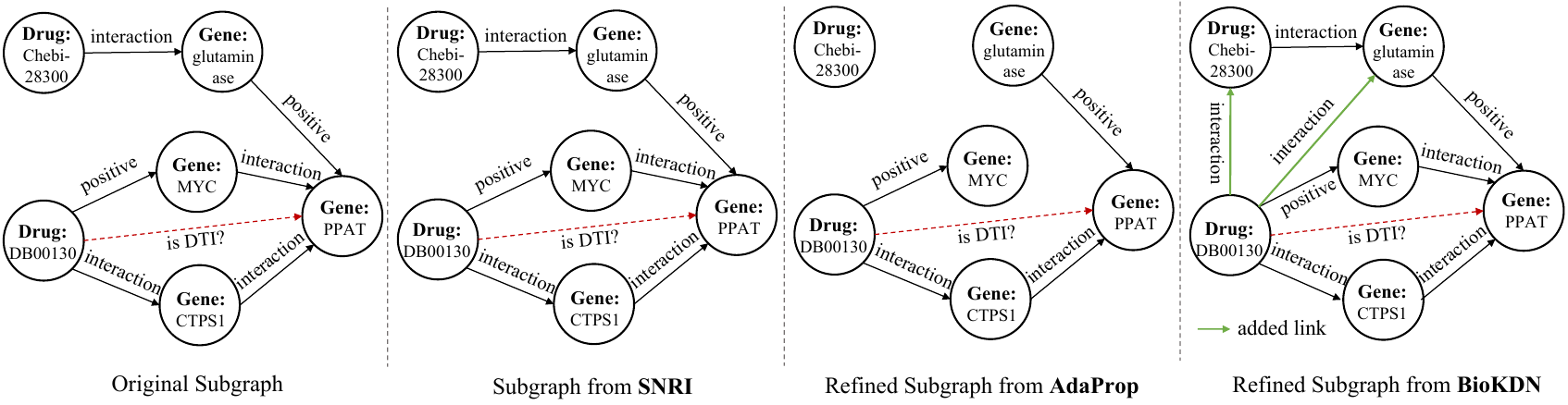} % Reduce the Fig. size so that it is slightly narrower than the column. Don't use precise values for Fig. width.This setup will avoid overfull boxes.
\caption{The original noisy subgraph and the refined subgraphs constructed from SNRI, AdaProp, and BioKDN.}
\label{fig_case}
\end{figure*}

\begin{figure}[t]
\centering
\includegraphics[width=1\columnwidth]{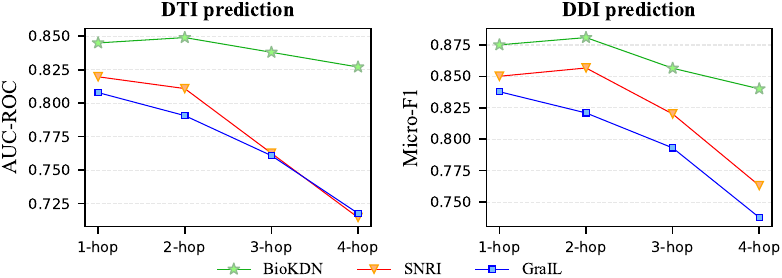}
\caption{The performance of BioKDN on Drugbank compared with SNRI and GraIL on various sizable subgraphs.}
\label{fig_k}
\end{figure}

\subsection{Robustness of Interaction Noises}
\subsubsection{Performance on Contaminated KGs}
To verify the effectiveness of structure reliability learning and smooth semantic preservation modules in denoising interaction noise, we generate different proportions of \textit{structural} and \textit{semantic} negative interactions (i.e., 25\%, 50\%, and 75\%) to contaminate the training knowledge graph. The reported performance of various models is evaluated on the unchanged test set shown in Fig.~\ref{fig_noise}. Structural noises are generated by sampling unknown triples from all possible \textit{entity-relation-entity} combinations, while semantic noises are sampled from all missing triples with reasonable \textit{entity-relation-entity} schemes (e.g., the scheme (drug, \textit{drug\_disease\_treat}, disease)). We then perform BioKDN and its variants (i.e., \textbf{ours w/o SSP} for structural noise and \textbf{ours w/o SRL} for semantic noise) on the noisy KG and compare their performance with semantic subgraph-based SNIR and global KG-based TriModel. 
As shown in Fig.~\ref{fig_noise} and Fig.~\ref{fig_noise_ddi}, the AUC-ROC and Micro-F1 values on DrugBank for DTI and DDI prediction tasks are reported. Additionally, the corresponding performance degradation \textcolor{black}{ratios} are presented. 

We observe that as more noise is added, the performance of all models deteriorates for both structural and semantic experiments. This is because the introduced noise weakens the expressive power of the aggregated neighbor information. However, BioKDN and its variants exhibit smaller degradation than other methods for both types of noise. The variants \textbf{ours w/o SSP} and \textbf{ours w/o SRL} show lighter changes as the noise increases than SNRI and TriModel on structural and semantic noise respectively, which indicates the 
structure reliability learning and smooth semantic subgraph can effectively denoise noisy interactions and ignore task-irrelevant relations. Furthermore, the gaps between BioKDN and SNRI or SumGNN grew larger with increasing noise. This is because BioKDN pays more attention to the informative interactions by maximizing the mutual information between the refined and semantic subgraphs.
This phenomenon shows that BioKDN can effectively mitigate noises using reliable structure and smoothed semantics.

\subsubsection{Impact of $k$-hop Subgraph}
To evaluate the global effectiveness of BioKDN, we conduct experiments to compare the impact of $k$-hop subgraphs with baselines SNRI~\cite{xu2022subgraph} and GraIL~\cite{teru2020inductive}. As shown in Fig.~\ref{fig_k}, BioKDN experiences less performance degradation than SNRI and GraIL in both DTI and DDI predictions. This occurs because as the subgraph size increases, more noise is introduced, which can negatively impact model performance. However, \textcolor{black}{the automatic denoising mechanism of BioKDN} helps minimize the adverse effects of noisy interactions. This finding demonstrates that BioKDN can effectively handle noise by leveraging a reliable structure and smooth semantics.

% To verify the effectiveness of BioKDN globally, we experiment to study the influence of k-hop subgraphs compared with baselines SNRI~\cite{xu2022subgraph} and GraIL~\cite{teru2020inductive}. In Fig., we observe that BioKDN has lighter degradation than SNRI and GraIL on both DTI and DDI predictions. This is because as the subgraph becomes larger, more noise is introduced which degrades the performance of the models, while BioKDN can reduce the negative impact of noisy interactions by the automatic denoising mechanism. This phenomenon shows that BioKDN can effectively mitigate noises using a reliable structure and smoothed semantics.

\subsection{Case Study}

We conduct a case study for predicting DTI relation between the drug \textit{DB00130} and the gene \textit{PPAT} to demonstrate the effectiveness of BioKDN, shown in Fig.~\ref{fig_case}. In the original subgraph, the different entities \textit{DB00130} and \textit{Chebi:28300} represent the same drug \textit{L-Glutamine}, but they are unalignment, resulting in the absence of an interactive edge (i.e., \textit{DB00130, same\_as, glutaminase}). By learning the reliable structure of the original subgraph, BioKDN effectively establishes a connection between the drug \textit{DB00130} and the gene \textit{glutaminase}, which brings favorable information for predicting the DTI relation between \textit{DB00130} and \textit{PPAT}. However, SNRI directly reasoning on the original subgraph without considering the noises. In addition, AdaProp uses a novel sample strategy to filter out task-irrelevant edges within the local subgraph and prune them, potentially ignoring reliable information and failing to build missing edges to bring information interaction.
This case shows that BioKDN can effectively learn a reliable structure, enhancing the performance of link prediction.

\section{Conclusion}
In this paper, we proposed a biomedical knowledge-enhanced denoising network, called BioKDN, to mitigate the negative influence caused by the noise and inconsistent semantics hiding in the KG. BioKDN designed both structure reliability learning and semantic smoothing modules to extract reliable structure and keep semantic relations consistent for robust DTI and DDI predictions. 
Our experiments on four datasets and contaminated KGs for DTI and DDI predictions demonstrate that BioKDN significantly outperforms several existing state-of-the-art methods. These results verify the robustness of our model against interaction noises. \textcolor{black}{However, the semantic smoothing strategy of BioKDN is limited when applied to other domains for denoising.}
In the future, we will design a generalized denoising model based on KGs to remove noisy facts in the real world rather than just being irrelevant to downstream tasks. \textcolor{black}{In addition, the effectiveness of reliable structure learning motivates us to design a robust relation discovery method for broader applications}. Based on this, we can construct a high-quality and reliable biomedical KG as the foundation for accelerating drug discovery.

\section*{Acknowledgement}
% The authors would like to thank colleagues and the anonymous reviewers who have provided valuable feedback to help improve the paper.
The work was supported by National Natural Science Foundation of China (Grant Nos. 62425204, 62122025, U22A2037, 62250028, 62272151, 61972138, 62106073, 62206089, 62202413, 62450002, 62432011), Hunan Provincial Natural Science Foundation of China (Grant Nos. 2024JJ4015, 2023JJ40178, 2022JJ20016), the science and technology innovation Program of Hunan Province (Grant Nos. 2022RC1099), Excellent Youth Funding of Hunan (Grant Nos. 23B0129), NSF under grants III-2106758, and POSE-2346158.

% {\appendix[Proof of the Zonklar Equations]
% Use $\backslash${\tt{appendix}} if you have a single appendix:
% Do not use $\backslash${\tt{section}} anymore after $\backslash${\tt{appendix}}, only $\backslash${\tt{section*}}.
% If you have multiple appendixes use $\backslash${\tt{appendices}} then use $\backslash${\tt{section}} to start each appendix.
% You must declare a $\backslash${\tt{section}} before using any $\backslash${\tt{subsection}} or using $\backslash${\tt{label}} ($\backslash${\tt{appendices}} by itself
%  starts a section numbered zero.)}

% %{\appendices
% %\section*{Proof of the First Zonklar Equation}
% %Appendix one text goes here.
% % You can choose not to have a title for an appendix if you want by leaving the argument blank
% %\section*{Proof of the Second Zonklar Equation}
% %Appendix two text goes here.}

% \section{References Section}
% You can use a bibliography generated by BibTeX as a .bbl file.
%  BibTeX documentation can be easily obtained at:
%  http://mirror.ctan.org/biblio/bibtex/contrib/doc/
%  The IEEEtran BibTeX style support page is:
%  http://www.michaelshell.org/tex/ieeetran/bibtex/
 
%  % argument is your BibTeX string definitions and bibliography database(s)
% %\bibliography{IEEEabrv,../bib/paper}
% You can manually copy in the resultant .bbl file and set second argument of $\backslash${\tt{begin}} to the number of references
%  (used to reserve space for the reference number labels box).

\bibliographystyle{IEEEtran}
\bibliography{main}

\begin{IEEEbiography}[{\includegraphics[width=1in,height=1.25in,clip,keepaspectratio]{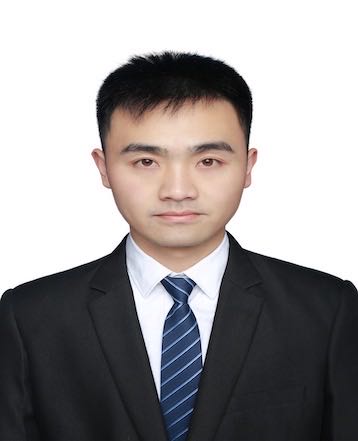}}]{Tengfei Ma}
received the Master's degree in computer science from Hunan University, Changsha, China, in 2022. He is currently a PhD student at the College of Computer Science and Electronic Engineering, Hunan University, Changsha, China. His main research interests include graph machine learning, knowledge graph representation, and their applications in science. He has published several research papers in these fields including \textit{IEEE Transactions on Knowledge and Data Engineering}, \textit{Bioinformatics}, \textit{Knowledge-Base Systems}, IJCAI, etc. 
\end{IEEEbiography}

\begin{IEEEbiography}
[{\includegraphics[width=1in,height=1.25in,clip,keepaspectratio]{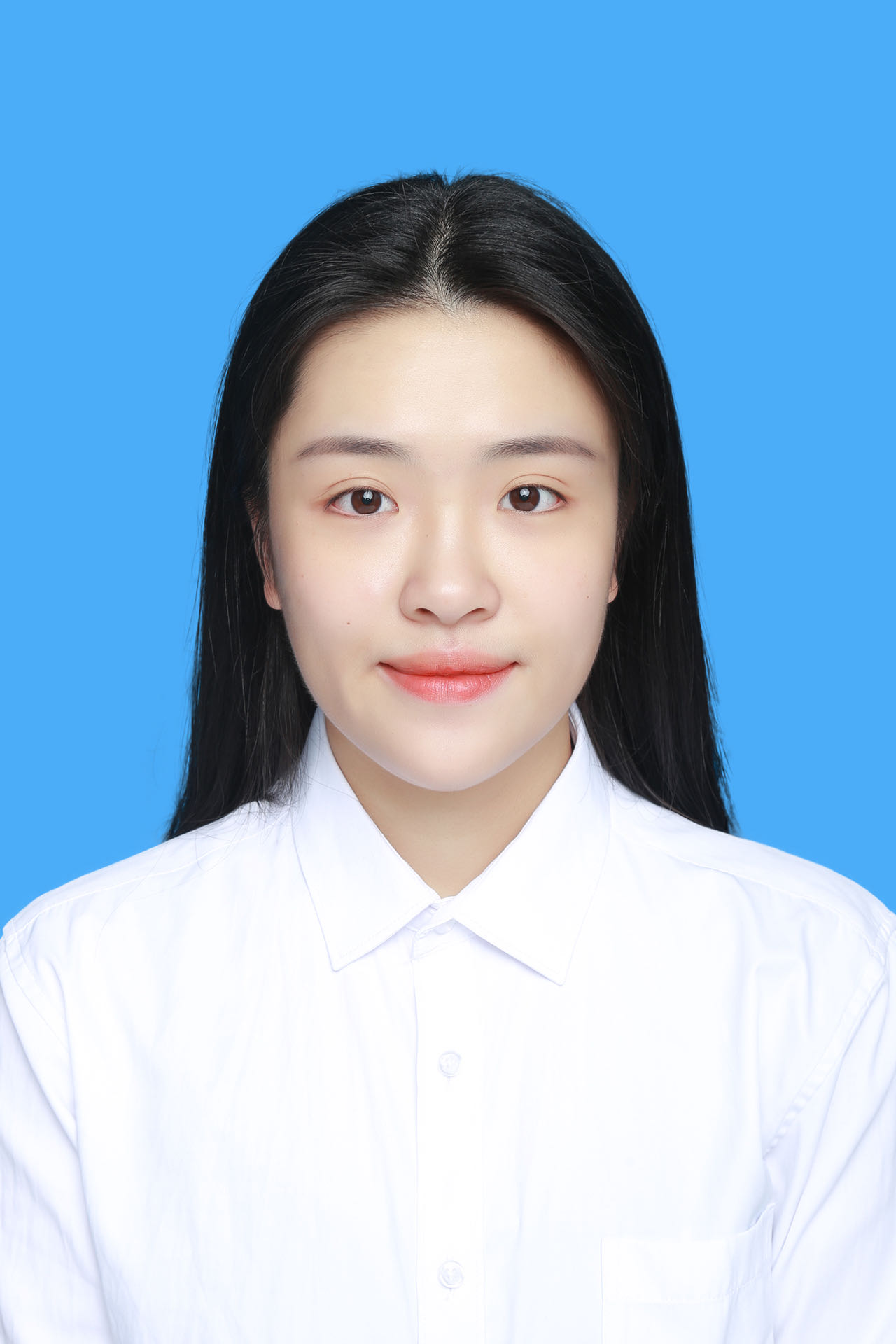}}]{Yujie Chen}
received her Master's degree in Computer Science from Hunan University, Changsha, China, in 2022. She is presently pursuing her PhD at the College of Computer Science and Electronic Engineering, also at Hunan University. Her research primarily focuses on knowledge graph representation, natural language processing, and graph machine learning, particularly their applications in the scientific domain. She has published a paper in this area, notably in the journal \textit{Bioinformatics}.
\end{IEEEbiography}

\begin{IEEEbiography}[{\includegraphics[width=1in,height=1.25in,clip,keepaspectratio]{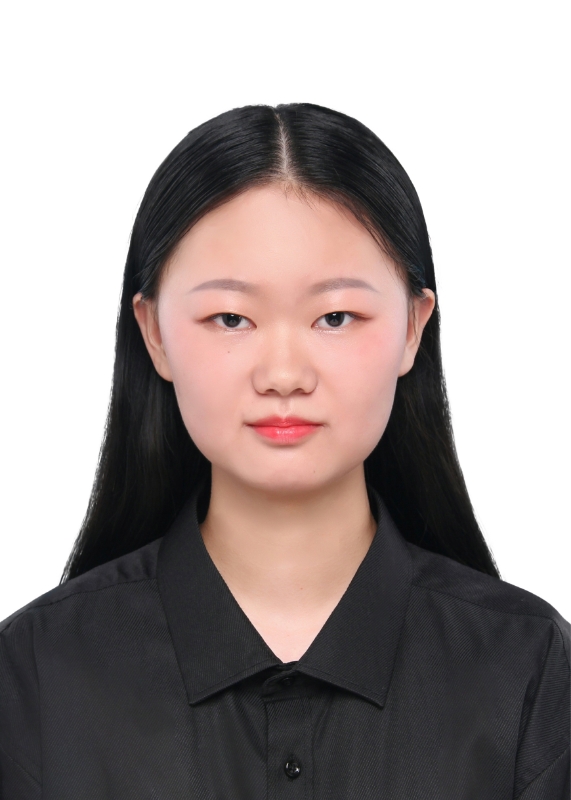}}]{Wen Tao}
received her M.E. degree in Computer Science and Technology from the College of Computer Science and Electronic Engineering, Hunan University, Changsha, China, in 2024. She received her B.E. degree in Computer Science and Technology from the School of Computer Science and Technology, China University of Mining and Technology, Xuzhou, China, in 2021. Her research interests include graph machine learning, drug discovery, and explainable AI.
\end{IEEEbiography}

\begin{IEEEbiography}[{\includegraphics[width=1in,height=1.25in,clip,keepaspectratio]{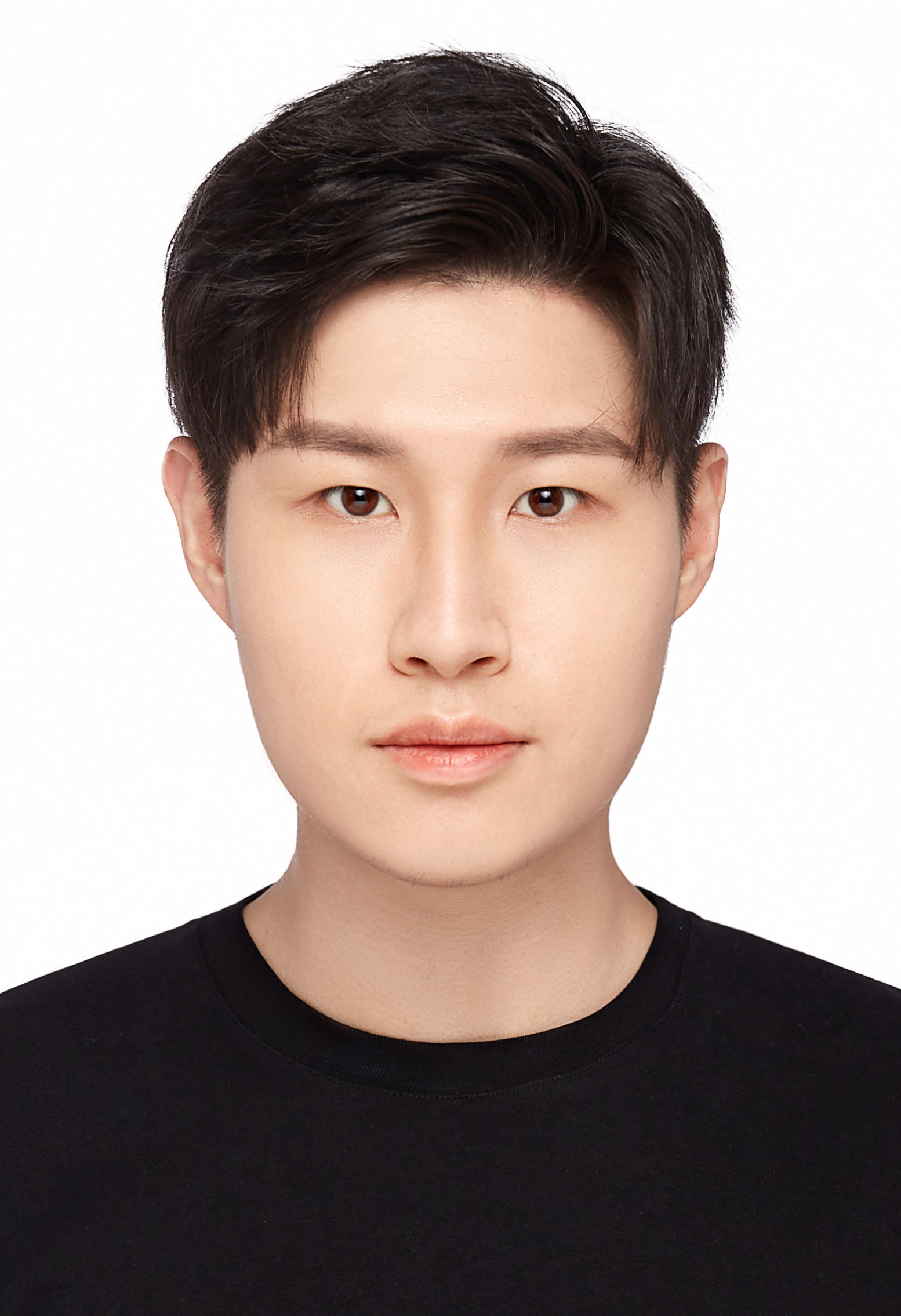}}]{Dashun Zheng}
obtained a Bachelor of Management degree in Information Management and Information Systems from Xinjiang University of Finance and Economics in 2022, and a Master's degree in Big Data and the Internet of Things from the Macao Polytechnic Institute in 2024. He is currently pursuing a Ph.D. in Computer Application Technology at the Macao Polytechnic Institute. His research interests include Natural Language Processing, AI for Healthcare, and Multimodal Learning.
\end{IEEEbiography}

\begin{IEEEbiography}
[{\includegraphics[width=1in,height=1.25in,clip,keepaspectratio]{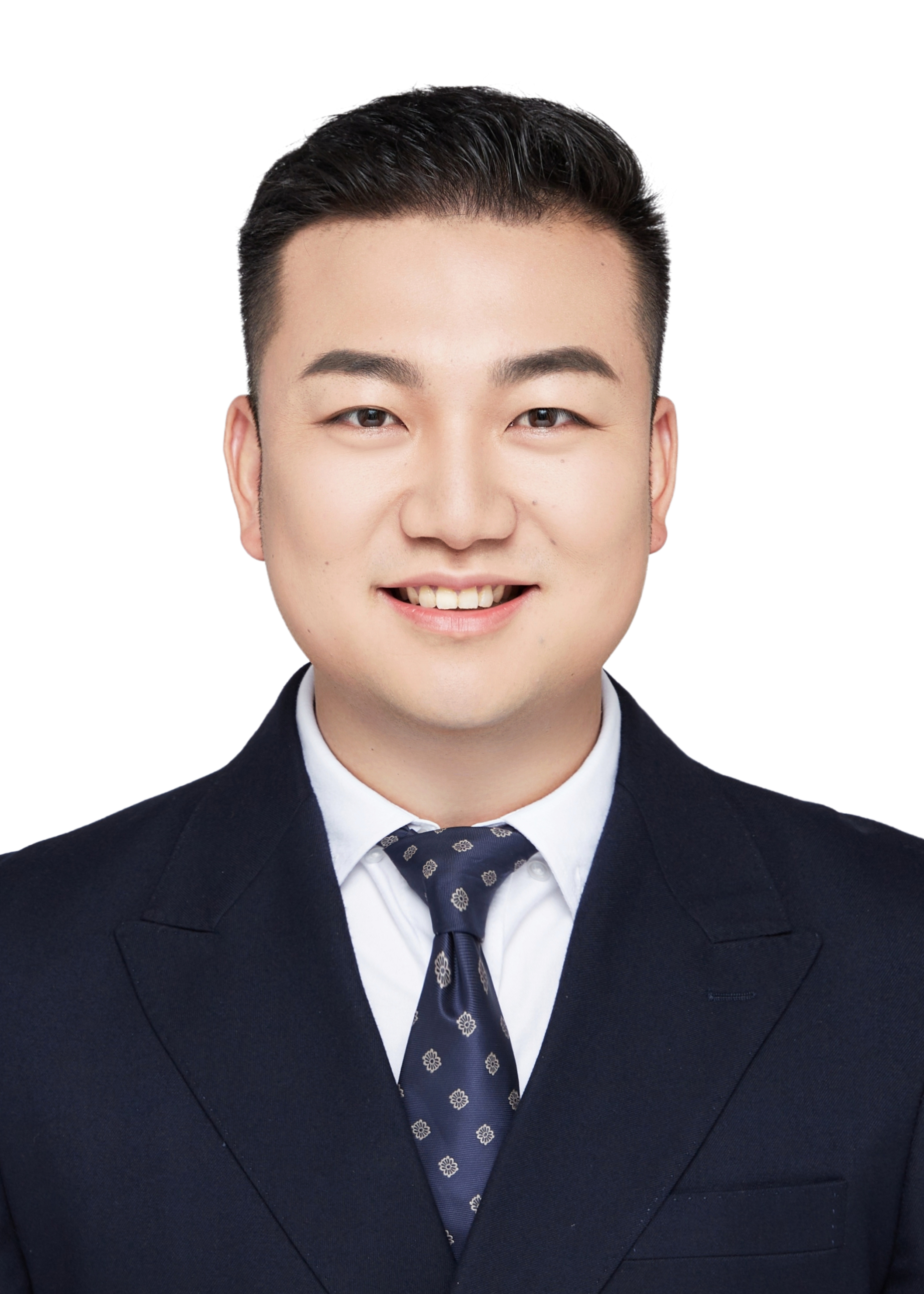}}]{Xuan Lin}
received the Ph.D. degree in computer science from Hunan University, Changsha, China, in 2021. He is currently a lecturer at the College
of Computer Science, Xiangtan University, Xiangtan, China. He was a visiting scholar with the University of Illinois at Chicago, from 2019 to 2020. His main research interests include machine learning, graph neural networks, and bioinformatics. He has published several research papers in
these fields including \textit{IEEE Transactions on Knowledge and Data Engineering}, \textit{Briefings in Bioinformatics}, IJCAI, AAAI, ECAI, BIBM, etc.
\end{IEEEbiography}

\begin{IEEEbiography}[{\includegraphics[width=1in,height=1.25in,clip,keepaspectratio]{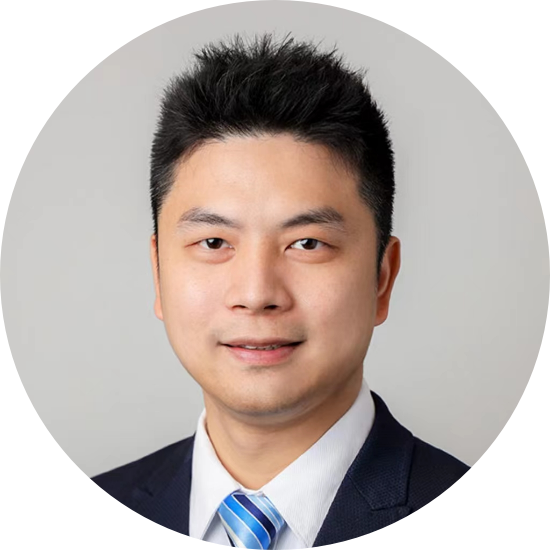}}]{Patrick Cheong-Iao Pang}
% % Use $\backslash${\tt{begin\{IEEEbiography\}}} and then for the 1st argument use $\backslash${\tt{includegraphics}} to declare and link the author photo.
is currently an Assistant Professor and a Ph.D. supervisor in the Faculty of Applied Sciences at Macao Polytechnic University (MPU). He received his Ph.D. from the University of Melbourne and held appointments at the University of Melbourne and Victoria University before joining MPU. His research interests include natural language processing (NLP) applications, and digital health and educational technologies. His interdisciplinary work seeks to build AI applications for health and education, thanks to his many years of industry experience in designing and building IT systems.
\end{IEEEbiography}

\begin{IEEEbiography}[{\includegraphics[width=1in,height=1.25in,clip,keepaspectratio]{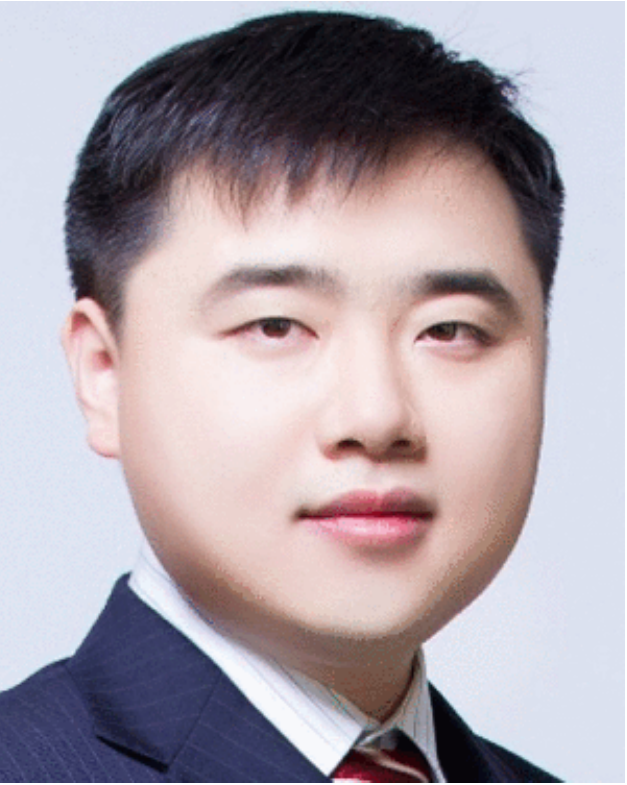}}]{Yiping Liu}
received a B.Eng. degree in electrical engineering and automation and a Ph.D. degree in control theory and control engineering from the China University of Mining and Technology, Xuzhou, China in 2012 and 2017, respectively. He is currently an Associate Professor at the College of Computer Science and Electronic Engineering, Hunan University, Changsha, China. During 2018–2020, he was a Research Assistant Professor at the Department of Computer Science and Intelligent Systems, at Osaka Prefecture University, Sakai, Japan. During 2016–2017, he was a Visiting Scholar with the School of Electrical and Computer Engineering, Oklahoma State University, Stillwater, OK, USA. His research interests include evolutionary computation, multi-objective optimization, and machine learning.
\end{IEEEbiography}

\begin{IEEEbiography}[{\includegraphics[width=1in,height=1.25in,clip,keepaspectratio]{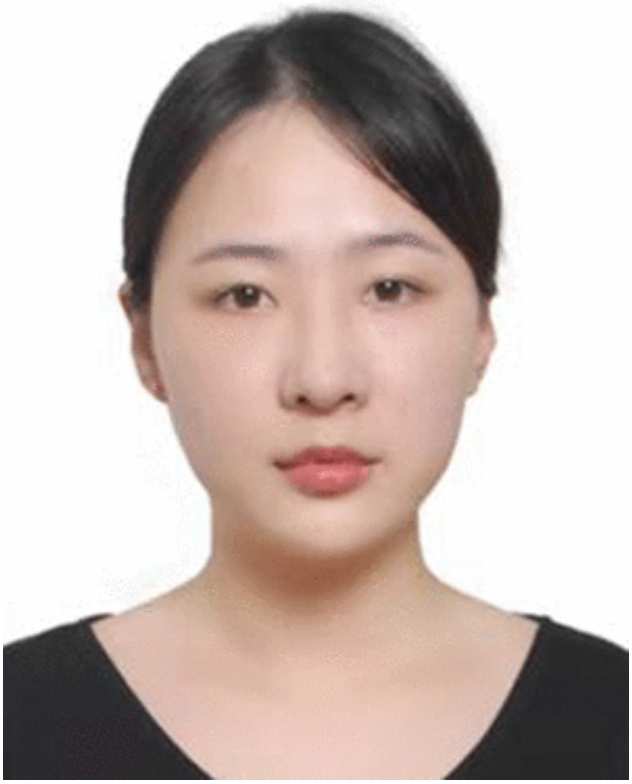}}]{Yijun Wang}
received the B.E. and Ph.D. degrees from the University of Science and Technology of China, Hefei, China, in 2014 and 2019, respectively. She is currently an Assistant Professor with the School of Computer Science and Electronic Engineering, at Hunan University, Changsha, China. She has authored or co-authored more than 10 papers in related conferences and journals. Her research interests include multimedia understanding, natural language processing, and data mining.
\end{IEEEbiography}

\begin{IEEEbiography}[{\includegraphics[width=1in,height=1.25in,clip,keepaspectratio]{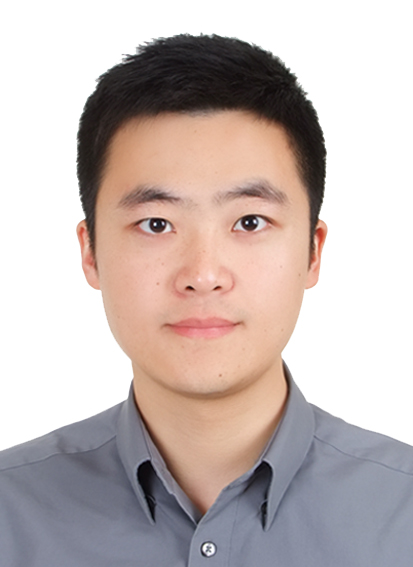}}]{Longyue Wang} (Senior Member, IEEE)
has received a B.S. degree in network engineering in 2011, and an M.S. degree in software engineering in 2014. From 2015 to 2018, he pursued a Ph.D. at Dublin City University. He is a senior research fellow at Tencent AI Lab now. He has studied and practiced in a broad field of Artificial Intelligence, especially in Large Language Models, Multimodal, Language Agents, Natural Language Processing, Machine Translation, Deep Learning, and AI for Science.
\end{IEEEbiography}

\begin{IEEEbiography}
[{\includegraphics[width=1in,height=1.25in,clip,keepaspectratio]{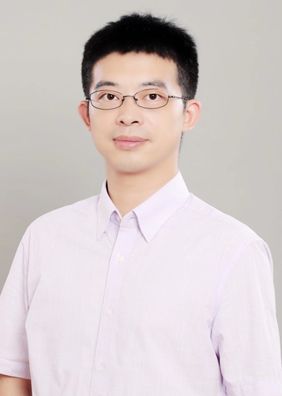}}]{Bosheng Song}
received a Ph.D. degree in control science and engineering from the Huazhong University of Science and Technology, Wuhan, China, in 2015. He spent eighteen months working with the Research Group on Natural Computing, University of Seville, Seville, Spain, from November 2013 to May 2015. He worked as a post-doctoral researcher with the School of Automation, Huazhong University of Science and Technology, from 2016 to 2019. He is currently an associate professor with the College of Information Science and Engineering, Hunan University, Changsha, China. His current research interests include membrane computing and bioinformatics.
\end{IEEEbiography}

\begin{IEEEbiography}
[{\includegraphics[width=1in,height=1.25in,clip,keepaspectratio]{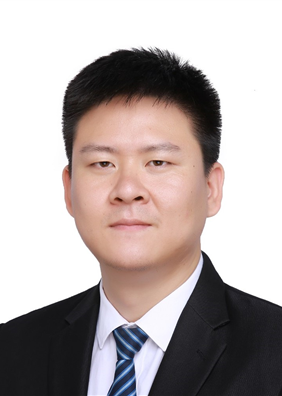}}]{Xiangxiang Zeng}
(Senior Member, IEEE) received a Ph.D. degree in system engineering from the Huazhong University of Science and Technology, China, in 2011. He is a Yuelu distinguished professor at the College of Information Science and Engineering, Hunan University, Changsha, China. Before joining Hunan University, in 2019, he was with the Department of Computer Science, Xiamen
University. He was a visiting scholar with Indiana University, The Chinese University of Hong Kong, Oklahoma State University, etc. His main research interests include computational intelligence, graph neural networks, and bioinformatics.
\end{IEEEbiography}

\begin{IEEEbiography}
[{\includegraphics[width=1in,height=1.25in,clip,keepaspectratio]{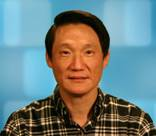}}]{Philip S. Yu}
\textcolor{black}{(Fellow, IEEE) received the B.S. degree in E.E. from National Taiwan University, the M.S. and Ph.D degrees in E.E. from Stanford University, and the M.B.A. degree from New York University. He is a Distinguished Professor in Computer Science at the University of Illinois Chicago and holds the Wexler Chair in Information Technology.   Before joining UIC, Dr. Yu was with IBM, where he was manager of the Software Tools and Techniques department at the Watson Research Center. His research interest is on big data, and artificial intelligence, including data mining, database and privacy. Dr. Yu is the recipient of ACM SIGKDD 2016 Innovation Award, the IEEE Computer Society’s 2013 Technical Achievement Award, and the Research Contributions Award from IEEE Intl. Conference on Data Mining (ICDM) in 2003. He also received the VLDB 2022 Test of Time Award, ACM SIGSPATIAL 2021 10-year Impact Award, WSDM 2020 Honorable Mentions of the Test of Time Award, ICDM 2013 10-year Highest-Impact Paper Award, and the EDBT Test of Time Award (2014). He was the Editor-in-Chiefs of ACM Transactions on Knowledge Discovery from Data (2011-2017) and IEEE Transactions on Knowledge and Data Engineering (2001-2004). He is a Fellow of the ACM.}

\end{IEEEbiography}

% % \bf{If you will not include a photo:}\vspace{-33pt}
% % Use $\backslash${\tt{begin\{IEEEbiographynophoto\}}} and the author name as the argument followed by the biography text.

% \vfill

\end{document}